\documentclass[11pt,twoside,reqno]{amsart}
\usepackage{amsmath, amsthm, amscd, amsfonts, amssymb, graphicx, color}
\usepackage[bookmarksnumbered, plainpages]{hyperref}

\usepackage{fancyhdr}
\usepackage{lastpage}
\usepackage{caption}
\usepackage{subcaption}
\usepackage{cite}
\usepackage[footskip=0.25in]{geometry}

\usepackage{courier}     

\usepackage{etoolbox}
\makeatletter
\def\@settitle{\begin{center}%
		\baselineskip14\p@\relax
		\bfseries
		\normalfont\LARGE
		\@title
	\end{center}%
}
\patchcmd{\@setauthors}{\MakeUppercase}{}{}{}
\def\section{\@startsection{section}{1}%
    \z@{.7\linespacing\@plus\linespacing}{.5\linespacing}%
    {\normalfont\bf}}
\def\subsection{\@startsection{subsection}{1}%
    \z@{\linespacing\@plus\linespacing}{.5\linespacing}%
    {\normalfont\it}}
\def\subsubsection{\@startsection{subsubsection}{1}%
    \z@{\linespacing\@plus\linespacing}{.5\linespacing}%
    {\normalfont\it}}
\makeatother

\newtheorem{theorem}{Theorem}[section]

\newtheorem{notation}{Notation}

\theoremstyle{definition}

\newtheorem{definition}[theorem]{Definition}

\makeatletter
\patchcmd{\@setauthors}{\centering}{\raggedright}{}{}
\patchcmd{\abstract}{3pc}{0pt}{}{} 
\makeatother

\usepackage{xargs} 
\usepackage{bbm} 
\usepackage{stackrel} 
\usepackage{xspace}  
\usepackage{diagbox} 
\usepackage{float}  
\usepackage{xpatch}
\makeatletter
\usepackage{framed}
\usepackage{xcolor}    
\usepackage{colortbl}  
\usepackage{listings} 
\definecolor{grigiochiarissimo}{rgb}{0.97,0.97,0.97}
\definecolor{grigiomoltochiaro}{rgb}{0.96,0.96,0.96}
\definecolor{grigiochiaro}{rgb}{0.8,0.8,0.8}
\definecolor{grigiomedio}{rgb}{0.6,0.6,0.6}
\definecolor{grigioscuro}{rgb}{0.3,0.3,0.3}
\definecolor{giallochiaro}{rgb}{1,1,0.95}
\definecolor{azzurrochiaro}[h=20]{rgb}{0.95,0.97,1}
\definecolor{blu}{rgb}{0,0,0.8}
\definecolor{bluscuro}{rgb}{0,0,0.5}
\definecolor{verdescuro}{rgb}{0,0.5,0}
\definecolor{arancione}{rgb}{1,0.5,0}
\definecolor{viola}{rgb}{0.65, 0.12, 0.82}
\definecolor{malva}{rgb}{0.58,0,0.82}
\definecolor{coloredisfondo}{rgb}{0.97,0.97,0.97}         
\definecolor{colorebordo}{HTML}{E0E0E0}                 
\definecolor{coloredibase}{HTML}{000000}                
\definecolor{colorenumerazione}{rgb}{0.8,0.8,0.8}       
\definecolor{coloreparolechiave}{rgb}{0,0,0.8}          
\definecolor{coloreparolechiave-2}{rgb}{0.65,0.12,0.82} 
\definecolor{coloreparolechiave-3}{HTML}{002DB2}        
\definecolor{coloreparolechiave-4}{HTML}{00238C}        
\definecolor{coloreparolechiave-5}{HTML}{00238C}        
\definecolor{colorestringhe}{rgb}{0,0.5,0}              
\definecolor{coloreidentificatori}{HTML}{000000}        
\definecolor{colorestringhe}{HTML}{067D17}              
\definecolor{coloredirettive}{HTML}{9D8740}             
\definecolor{colorenumero}{HTML}{001A66}                
\definecolor{coloredelimitatori}{HTML}{FF0000}
\definecolor{coloredelimitatoriangolari}{rgb}{0.65, 0.12, 0.82}  
\definecolor{coloredelimitatoriquadre}{rgb}{0.65, 0.12, 0.82}    
\definecolor{colorecommenti}{rgb}{0.6,0.6,0.6}                   
\definecolor{colorecommenti-2}{HTML}{4497E7}                     
\lstdefinestyle{stilePython}{
language=Python,%
keywordstyle={\bfseries\color{coloreparolechiave}},            
morekeywords={},          
keywords=[2]{self,len},                                        
keywordstyle={[2]{\color{coloreparolechiave-2}}},              
keywords=[3]{},                                                
keywordstyle={[3]{\color{coloreparolechiave-3}}},              
keywords=[4]{},                                                
keywordstyle={[4]{\color{coloreparolechiave-4}}},              
keywords=[5]{},                                                
keywordstyle={[5]{\color{coloreparolechiave-5}}},              
identifierstyle=\color{coloreidentificatori},                  
stringstyle={\ttfamily \color{colorestringhe}},                
commentstyle=\color{colorecommenti},                           
morecomment=[s][\color{colorecommenti-2}]{/*}{*/},             
literate=%
{0}{{\color{colorenumero}{0}}}{1}%
{1}{{\color{colorenumero}{1}}}{1}%
{2}{{\color{colorenumero}{2}}}{1}%
{3}{{\color{colorenumero}{3}}}{1}%
{4}{{\color{colorenumero}{4}}}{1}%
{5}{{\color{colorenumero}{5}}}{1}%
{6}{{\color{colorenumero}{6}}}{1}%
{7}{{\color{colorenumero}{7}}}{1}%
{8}{{\color{colorenumero}{8}}}{1}%
{9}{{\color{colorenumero}{9}}}{1},                 
backgroundcolor=\color{coloredisfondo},                        
frame=none,                                                      
xleftmargin=8pt,  
xrightmargin=4pt,  
framesep=2pt,                                                  
framerule=0.6pt,                                                 
rulecolor=\color{colorebordo},                                 
numbers=left,                                                  
numberstyle=\tiny\color{colorenumerazione},                    
numbersep=3pt,                                                 
numberblanklines=false,                                        
showlines=false,                                               
firstnumber=1,                                                 
breakatwhitespace=false,                                       
breaklines=true,                                               
captionpos=b,                                                  
keepspaces=true,                                               
showspaces=false,                                              
showstringspaces=false,                                        
showtabs=false,                                                
columns=fixed,                                                 
tabsize=3,                                                     
extendedchars=true,                                            
escapeinside=$$                     
}
\lstdefinestyle{stilePythonambiente}{
style=stilePython,
basicstyle={\footnotesize\ttfamily\color{coloredibase}},
firstnumber=1,
lineskip=-1pt
}
\lstdefinestyle{stilesemplicecomando}{
basicstyle={\ttfamily\color{coloredibase}},
mathescape
}
\lstdefinestyle{stilePythoncomando}{
style=stilePython,
basicstyle={\ttfamily\color{coloredibase}},
mathescape
}
\lstdefinestyle{stilePythonconsole}{
style=stilePython,
numbers=none,
backgroundcolor=\color{giallochiaro},
basicstyle={\footnotesize\ttfamily\color{coloredibase}},
lineskip=-2pt,
otherkeywords={>>>}
}
\newcommand{\cpy}[1]{\lstinline[style=stilePythoncomando]!#1!}    
\newcommand{\cpysimple}[1]{\lstinline[style=stilesemplicecomando]!#1!}    

\lstnewenvironment{codpy}{%
\lstset{style=stilePythonambiente}}{}
\lstnewenvironment{codpyconsole}{%
\lstset{style=stilePythonconsole}}{}

\usepackage{pgf-umlcd}   

\tikzset{
  reduce height/.style={
    minimum height=0pt,
    inner ysep=0pt,
    text depth=2pt
  },
  reduce height/.default={0pt}
}

\newcommand{\df}[1]{\textit{#1}}  
\newcommand{\U}{\mathbbmss{U}}
\newcommand{\V}{\mathbbmss{V}}
\newcommand{\RR}{\mathbb{R}} 

\newcommand{\und}{\underline}  

\newcommand{\M}[1]{\mu_{#1}}       
\newcommand{\I}[1]{\sigma_{#1}}       
\newcommand{\NM}[1]{\omega_{#1}}   
\newcommand{\DM}[2][u]{\mu_{#2}\left(#1\right)}       
\newcommand{\DI}[2][u]{\sigma_{#2}\left(#1\right)}       
\newcommand{\DNM}[2][u]{\omega_{#2}\left(#1\right)}   
\newcommand{\ANG}[1]{\left\langle #1 \right\rangle}  
\newcommand{\NSbase}[4][\U]{\ANG{#1, #2, #3, #4}}  
\newcommand{\ns}[1]{\widetilde{#1}}   
\newcommand{\NS}[2][\U]{\NSbase[#1]{\M{#2}}{\I{#2}}{\NM{#2}}}   
\newcommand{\nNS}[2][\U]{\ns{#2} = \NS[#1]{#2}}  
\newcommandx{\NSEXT}[3][1=u,2=\U]{\left\{ \left( #1, \DM{#3}, \DI{#3}, \DNM{#3} \right) : \, #1 \in #2 \right\}} 
\newcommand{\nameSVNS}[1][]{SVN-set#1\xspace}  
\newcommand{\NSemptyset}[1][]{\widetilde{\emptyset}_{#1}}
\newcommand{\NSabsoluteset}[1][\U]{\widetilde{#1}}
\newcommand{\NSsubseteq}{\mathrel{\ooalign{
\raise0.2ex\hbox{$\Subset$}%
\cr\hidewidth%
\raise-0.25ex\hbox{\rule[0pt]{5.6pt}{0.4pt}}%
\hidewidth\cr%
}}}
\newcommand{\NSsupseteq}{\mathrel{\ooalign{
\raise0.2ex\hbox{$\Supset$}%
\cr\hidewidth%
\raise-0.25ex\hbox{\rule[0pt]{5.6pt}{0.4pt}}%
\hidewidth\cr%
}}}
\newcommand{\NSeq}{\mathrel{\ooalign{
\raise0.1ex\hbox{$=$}%
\cr%
\hskip0.6pt\raise0.2ex\hbox{\rule{6.6pt}{0.35pt}}
\cr\hidewidth%
\raise1.05ex\hbox{{\rule{6.6pt}{0.32pt}}}
\hidewidth\cr%
}}}
\newcommand{\NSneq}{\not\NSeq}   
\newcommand{\NScompl}{{\hskip.15ex\ooalign{\hbox{\scalebox{0.8}{$\complement$}}%
\hidewidth\cr\hspace{1.6pt}\hbox{\rule[0.2pt]{0.6pt}{6.6pt}}%
}}}    
\newcommand{\NScup}{\Cup}  
\newcommand{\NScap}{\Cap}  
\newcommand{\NSsetminus}{\hspace{-0.3em}\setminus\hspace{-0.88em}\setminus\hspace{0.1em}}   
\newcommand{\inv}[1]{{#1}^{-1}\!}  
\newcommand{\fibre}[2]{{#1}^{-1}\left(\left\{#2\right\}\right)}    

\newcommand{\dunder}{\ensuremath{\_\hspace{0.2pt}\_}}   

{\begin{enumerate}[label={\rm(\arabic*)},topsep=4pt]%
\itemsep1pt
\setcounter{enumi}{#1}%
}%
{\end{enumerate}}

{\begin{enumerate}[label={\rm(\alph*)},topsep=4pt]%
\itemsep1pt
\setcounter{enumi}{#1}%
}%
{\end{enumerate}}

{\begin{enumerate}[label={\rm(\roman*)},topsep=4pt]%
\itemsep1pt
\setcounter{enumi}{#1}%
}%
{\end{enumerate}}

{\begin{tabular}[t]{|c|c|c|c|}\hline
\diagbox{\quad$#1$}{\\[1mm]\quad$#2$} &
$#3$ & $#4$ & $#5$ \\ \hline}%
{\\ \hline \end{tabular}}

\usepackage{dirtree}   
\DTsetlength{0.2em}{20pt}{0.2em}{0.8pt}{4pt}

\usepackage{setspace}  

\usepackage[ruled,vlined]{algorithm2e}  
\SetAlgoNlRelativeSize{-1}
\newenvironment{algoritmo}{%
\begin{algorithm}[H]%
\setstretch{0.9}%
\NoCaptionOfAlgo  
\LinesNotNumbered 
\DontPrintSemicolon 
}
{\end{algorithm}}

\begin{document}

\tolerance=10000 \baselineskip18truept
\newbox\thebox
\global\setbox\thebox=\vbox to 0.2truecm{\hsize
0.15truecm\noindent\hfill}
\def\boxit#1{\vbox{\hrule\hbox{\vrule\kern0pt
     \vbox{\kern0pt#1\kern0pt}\kern0pt\vrule}\hrule}}
\def\qed{\lower0.1cm\hbox{\noindent \boxit{\copy\thebox}}\bigskip}
\def\ss{\smallskip}
\def\ms{\medskip}
\def\nt{\noindent}
\renewcommand\footnoterule{\hskip-10pt\rule{\linewidth}{0.05pt}}
\let\thefootnote\relax

\title{A Python Framework for Neutrosophic Sets and Mappings}
\author{Giorgio Nordo$^{1,*}$, Saeid Jafari$^2$, Arif Mehmood$^3$ and Bhimraj Basumatary$^4$\\
    \footnotesize{$^{1}$MIFT Department -- Mathematical and Computer Science, Physical Sciences and Earth Sciences\\
University of Messina, 98166 Sant'Agata, Messina, Italy; giorgio.nordo@unime.it}\\
    \footnotesize{$^{2}$Mathematical and Physical Science Foundation, 4200 Slagelse, Denmark; jafaripersia@gmail.com}\\
    \footnotesize{$^{3}$Department of Mathematics, Institute of Numerical Sciences, Gomal University, Dera Ismail Khan 29050, KPK, Pakistan; mehdaniyal@gmail.com}\\
    \footnotesize{$^{4}$Department of Mathematical Sciences, Bodoland University, Kokrajhar, BTAD, India;
    brbasumatary14@gmail.com}\\
    \footnotesize{$^{*}$Correspondence: giorgio.nordo@unime.it}%
\footnote{\hskip-12pt{Nordo G., Jafari S., Mehmood A., Basumatary B., A Python Framework for Neutrosophic Sets and Mappings}}
}

\maketitle

\renewcommand\abstractname{\textbf{Abstract}}
\begin{abstract}
In this paper we present an open source framework developed in Python and consisting of three distinct
classes designed to manipulate in a simple and intuitive way
both symbolic representations of neutrosophic sets over universes of various types
as well as mappings between them.
The capabilities offered by this framework extend and generalize
previous attempts to provide software solutions to the manipulation of neutrosophic sets
such as those proposed by Salama et al. \cite{salama2014c},
Saranya et al. \cite{saranya2020},
El-Ghareeb \cite{el-ghareeb2019},
Topal et al. \cite{topal2019} and Sleem \cite{sleem2020}.
The code is described in detail and many examples and use cases are also provided.
{\bf
-----------------------------------------------------------------------------------------------------------------------------}
\end{abstract}
\thispagestyle{empty}


\section{Introduction}
Since the notion of neutrosophic set was introduced in 1999 by Smarandache \cite{smarandache}
as a generalization of both the notions of fuzzy set introduced by Zadeh\cite{zadeh} in 1965 and
intuitionistic fuzzy set proposed by Atanassov \cite{atanassov} in 1983,
neutrosophic set theory had a rapid development
and has been profitably used in many fields of pure Mathematics \cite{abobala,mehmood2020,mehmood2020a,nordo2020,saber2020}
as well as in several areas of applied sciences such as
Graph Theory \cite{broumi},
Decision Making \cite{mondal},
Medicine \cite{broumi2022},
Statistics \cite{smarandache2014,kunwar,schweizer},
Image Analysis \cite{guo,zhang}
Machine Learning \cite{elhassouny,sharma}, etc.

In numerous instances, especially when dealing with applications stemming from real-world issues,
manually manipulating neutrosophic sets that possess a finite yet consistent number of elements,
along with their associated mappings, can be quite laborious and challenging.
Consequently, there exists a significant demand for a system that can streamline
the automation of key neutrosophic operations, including union, intersection, neutrosophic difference,
and the calculation of neutrosophic images or counterimages by mappings.
Previous attempts to address this need were undertaken by Salama et al., who initially
employed tools like Microsoft Excel \cite{salama2014b} and later transitioned
to using C\# \cite{salama2014c}.
Another software application for processing neutrosophic sets developed in C\#
was described by Saranya et al. \cite{saranya2020}.
More recently, El-Ghareeb introduced a Python package designed to handle both single
and interval-valued neutrosophic numbers and sets \cite{el-ghareeb2019}.
Unfortunately, however, in the latter paper the two classes concerning neutrosophic sets are described
incompletely and summarily than those concerning neutrosophic numbers.
Version 0.0.5 of this software that we consulted does not appear to provide adequate functionality
even for the main neutrosophic operations
and in any case the related repository on GitHub of the source code mentioned in the article
does not appear to be available.
Furthermore, other Python-based software solutions for handling neutrosophic numbers
and matrices have been proposed by Topal et al. \cite{topal2019} and Sleem \cite{sleem2020}.
However, the authors are not aware of any other Python software specifically designed
for the manipulation of neutrosophic sets is currently known.

This underscores the ongoing requirement for a set of well-structured Python classes,
ideally available under an Open Source license, that enable automated and interactive manipulation
of symbolic representations of neutrosophic sets, along with their associated mappings.
Additionally, there is a need for comprehensive documentation and user-friendly design
to facilitate straightforward integration for future implementations.


For this reason, we intended to design and develop a modern framework
that extends and generalizes the above software solutions
overcoming some of their limitations and offering greater flexibility in their use,
including interactive, aimed at the manipulation of neutrosophic sets and functions.
The structure of the entire framework has been carefully described by means of
of Unified Modeling Language (UML for short), a modeling and specification description language
very popular in Software Engineering.
The underlying given structures as well as the most significant methods of each of the classes of which the
the framework have been explained in detail in order to allow for further future refinements
of both a theoretical and applicative nature.

We are confident that the necessity mentioned has been effectively tackled through
the Python framework outlined in this paper.
The complete source code for this framework has been released under
the Open Source GNU General Public License version 3.0 (or GPL-3.0)
and is freely accessible at the url
\href{https://github.com/giorgionordo/pythonNeutrosophicSets}{github.com/giorgionordo/pythonNeutrosophicSets}.

In particular, Section 2 describes the general structure of the framework,
the dependency relationships among the various classes that comprise it as well as the reasons that
suggested the use of the Puython language and the decision to release the entire code produced
under an Open Source license.

Section 3 introduces some useful notations for extending some data structures
typical of Python in order to create a flexible substandard for describing neutrosophic sets
and functions between them.
In addition, some utility functions are briefly called that later will be invoked by some classes in the framework.

Section 4 contains a description of the properties and methods of the class \cpy{NSuniverse}
used to represent the universe sets on which the neutrosophic sets will be defined.

Section 5 describes in great detail
the properties and methods that make up the \cpy{NSset} class,
the main class of the framework, used for the representation of neutrosophic sets
and, in addition, numerous examples of practical use are also provided both
in both traditional and interactive environments.

Section 6 is devoted to the description of the class \cpy{NSmapping}
by which functions between two neutrosophic sets are represented.
The properties and methods of this class are described in detail
and illustrated with several practical examples.

Finally, some final remarks are made in Section 7,
highlighting the strengths of the framework presented in this paper
and inviting other researchers to continue, extend and improve
the development of the code described here.


\section{The framework \cpy{PYNS}}
The PYthon Neutrosophic Sets framework (\cpy{PYNS} for short) described
in the present paper consists of three classes designed to manage  respectively universe sets
(the \cpy{NSuniverse} class), neutrosophic sets (the \cpy{NSset} class)
and functions between them (the \cpy{NSmapping} class.
As is natural, the class \cpy{NSset} depends on (i.e. uses) the class \cpy{NSuniverse} class)
while the \cpy{NSmapping} class uses the other two as described in the following UML diagram of classes.
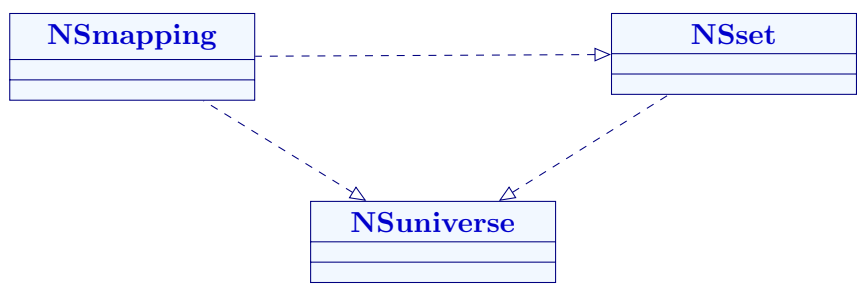
\vspace{2mm}
\begin{center}
\begin{tikzpicture}

  \begin{class}[text width=3cm]{NSuniverse}{4,0}
  \end{class}

  \begin{class}[text width=3cm]{NSset}{8,2.5}
    \implement{NSuniverse}
  \end{class}

  \begin{class}[text width=3cm]{NSmapping}{0,2.5}
    \implement{NSuniverse}
    \implement{NSset}
  \end{class}

\end{tikzpicture}
\end{center}
where the dashed arrow means "uses".


These three classes are respectively contained in the Python files
\cpy{ns_universe.py}, \cpy{ns_set.py} and \cpy{ns_mapping.py} which are located
in the package directory \cpy{pyns}.
The same directory contains also the file \cpy{ns_util.py} where are defined some
utility functions external to the classes but employed by them.
The structure of the package is described by the following diagram.
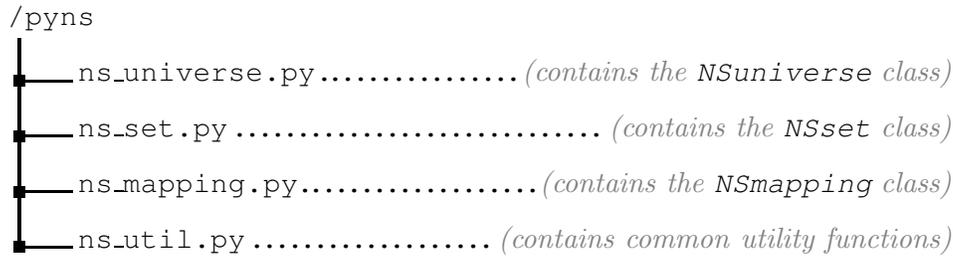
\vspace{-2mm}
\begin{center}
\begin{minipage}{0.85\textwidth} 
\dirtree{%
.1 /pyns.
.2 ns\_universe.py\DTcomment{(contains the \cpy{NSuniverse} class)}.
.2 ns\_set.py\DTcomment{(contains the \cpy{NSset} class)}.
.2 ns\_mapping.py\DTcomment{(contains the \cpy{NSmapping} class)}.
.2 ns\_util.py\DTcomment{(contains common utility functions)}.
}
\end{minipage}
\end{center}
\vspace{2mm}


The choice of programming language is a crucial aspect in the development of a scientific framework,
since it determines the performance, flexibility and ease of use of the entire system.
In the specific case of our framework, the choice to implement it using the Python language is based on several factors:
\begin{itemize}
\item Clear and expressive syntax: Python is known for its simple and readable syntax,
which makes the code more intuitive to write and understand.
This feature is especially relevant for a scientific framework,
as it facilitates the creation and manipulation of complex data structures such as neutrosophic sets and mappings between them.
\item Extensive standard library: Python offers an extensive standard library covering multiple scientific and mathematical domains.
This allows developers to easily use existing functions and tools to implement complex algorithms and optimize the performance of the framework.
\item Easy integration with other technologies: Python is known for its ability to integrate
with other programming languages and external libraries.
This is particularly useful in a scientific context, where it may be necessary to use
specialized libraries or existing computational tools.
\item Quick learning: Python is often considered one of the most accessible languages
even for programming novices.
Its relatively smooth learning curve allows students, researchers
and less experienced developers to tackle the framework with greater ease,
thus encouraging its dissemination and adoption in the scientific domain.
\end{itemize}

In summary, the choice of Python as the main language for the \cpy{PYNS} framework was crucial
in making the entire system more accessible, flexible, and powerful.
It allowed developers to focus more on scientific challenges
and mathematics specifics, rather than the complexities of the programming language,
thus accelerating the development and adoption of this important research tool.


Just as the Python language is released under an open source license approved by the Open Source Initiative (OSI),
the framework described in this paper is also made available as open code.
This offers numerous benefits and incentives for both developers and the scientific community as a whole, including:
\begin{itemize}
\item Knowledge sharing: The release of the framework under an open source license promotes
the sharing of knowledge and scientific discoveries.
By allowing anyone to access the source code, developers and researchers can
learn from others, build on others' work, and contribute improvements and new ideas.
\item Collaboration: The open source license encourages collaboration among experts and researchers from
from different academic institutions, organizations and countries.
This synergy can lead to faster developments, new discoveries, and innovative solutions to complex problems.
\item Transparency and verifiability: The availability of source code allows for
greater transparency in the implementation of the framework.
The scientific community can verify and validate the results obtained, increasing confidence
towards the framework and the results obtained through it.
\item Adaptability and customization: Users can adapt the framework to their own
specifics and customize it to address unique problems.
This flexibility results in a greater number of possible applications and uses
of the framework in various scientific contexts.
\item Cost reduction: Releasing the framework as open source eliminates the costs associated
of purchasing licenses or copyrights.
This allows academic institutions and organizations with limited resources to
free access to advanced scientific analysis tools and uncertain data.
\item Community growth: The adoption of the open source license attracts
a community of developers, researchers and enthusiasts interested in the field of
neutrosophic sets that can contribute to the evolution of the framework by providing feedback,
reporting bugs, and participating in the development of new features.
\item Continuity and longevity: The open source model can ensure greater
longevity of the framework, as it is not dependent on a single developer or institution.
The community can take care of the project over time, ensuring that it is always
updated and supported, even if there are changes in the original organization.
\end{itemize}


In particular, our framework \cpy{PYNS} is available under GNU General Public License version 3.0 (or GPL-3.0),
a generic software license developed by the Free Software Foundation (FSF) that provides users with
a set of rights and freedoms to use, modify, and distribute the software covered by the license.
More specifically, GPL-3.0 allows:
\begin{itemize}
\item Freedom of Use: the software may be used for any purpose, whether personal or commercial.
\item
Freedom to Study and Modify: you can analyze and study the source code of the software to understand how it works
and make changes to it according to your own needs.
\item Freedom of Distribution: you can distribute copies of modified or unmodified software to anyone,
while complying with the requirements of GPL-3.0.
\item Sharing of Changes: if you distribute modified software, you have an obligation to make available
the source code of your changes as well.
\item Compatibility with Derivative Works (copyleft): any derivative work based on software covered by GPL-3.0
must also be released under GPL-3.0 or a compatible license.
\end{itemize}


In conclusion, from the evaluation of all these aspects, it follows that
releasing the \cpy{PYNS}framework under GPL-3.0 license represents
a strategic choice that promotes innovation, collaboration and the dissemination
of knowledge in the field of neutrosophic theory,
thus promoting the advancement of scientific research in this area.


\section{Conventions and utility functions}
In the following we will make extensive use of Python's \cpy{dict} (dictionary) data structure
both for the internal representation of neutrosophic sets and for the definition of functions
between universe sets.
To make it even easier and more streamlined to use such structures both in interactive mode
as well as in writing client code based on such classes, it was chosen to also allow their representation
as a string and in free format, i.e., leaving the user free to:
\begin{itemize}
\item indifferently use not only the usual symbol \cpy{:} (colon)
but also alternatively the strings \cpy{->} (arrow) and \cpy{|->} (maps-to)
as separators between keys and values
\item indifferently use not only the usual symbol \cpy{,} (comma) but also \cpy{;} (semicolon) as
separators of the value-key pairs
\end{itemize}
in any combination thereof, and we will refer to this type of representation by the name \df{extended dictionary}.
In other words, while a classical Python dictionary has a form like:
$$\left\{ \textit{key}_1:\textit{value}_1, \textit{key}_2:\textit{value}_2,
\ldots \textit{key}_n:\textit{value}_n \right\} \, ,$$
an extended dictionary can be expressed as strings of the type:
$$\texttt{"} \textit{key}_1 \texttt{->} \textit{value}_1 \texttt{,}
\textit{key}_2 \texttt{|->}  \textit{value}_2 \texttt{;}
\ldots \textit{key}_n \texttt{->}  \textit{value}_n \texttt{"} \, .$$

The already mentioned file \cpy{NS_util.py} contains some general utility functions
that will be used repeatedly in the classes we will describe later.
More specifically, these functions are:
\begin{itemize}
\item \cpy{NSreplace($\textit{\textrm{text}}$, $\textit{\textrm{sostituz}}$)} which performs a series
of substitutions on the string \textit{text}
by replacing each key in the \textit{sostituz} dictionary with its corresponding value;
in particular, if that value is the null string \cpy{""} the effect will be
to remove all occurrences of the key,
\item \cpy{NSstringToTriplesList($\textit{\textrm{text}}$)}
which converts the string \textit{text} containing a list of triples
into the corresponding data structure by using the function \cpy{findall}
contained in the module \cpy{re} (regular expression)
and the function \cpy{literal_eval} contained in the module \cpy{ast} (Abstract Syntax Trees)
which allows interpreting data expressions contained in a string,
\item \cpy{NSisExtDict($\textit{\textrm{obj}}$)} that checks whether the object \textit{obj} passed as parameter
is a string representing an extended dictionary and returns the Boolean value \cpy{True} if it is,
\item \cpy{NSstringToDict($\textit{\textrm{text}}$)} which converts the string \textit{text} containing
an extended dictionary in a real Python dictionary,
\item \cpy{NSsplitText($\textit{\textrm{text}}$, $\textit{\textrm{max\_length}}$)} which returns the string \textit{text}
splitted into multiple lines of length not exceeding the value \textit{max\_length}.
\end{itemize}


The complete code for these functions is given in the following listing.

\begin{framed}
\begin{codpy}
from re import findall
from ast import literal_eval

def NSreplace(text, sostituz):
    for k in sostituz:
        text = text.replace(k, sostituz[k])
    return text

def NSstringToTriplesList(text):
    pattern = r'\[.*?\]|\(.*?\)'
    str_list = findall(pattern, text)
    tpl_list = [tuple(literal_eval(s)) for s in str_list]
    return tpl_list

def NSisExtDict(obj):
    result = False
    if type(obj) == str:
        result = (":" in obj) or ("->" in obj)

def NSstringToDict(text):
    sostituz = {"'": "", '"': "", "(": "", ")": "", "[": "", "]": "",
                "{": "", "}": "", " ": ",", ";": ",", ",,": ",",
                "|->": ":", "->": ':'}
    text = NSreplace(text, sostituz)
    listcouples = text.split(',')
    diz = dict()
    for couple in listcouples:
        key, value = couple.split(':')
        diz[key] = value
    return diz

def NSsplitText(text, max_length):
    words = text.split()
    lines = []
    current_line = ""
    for word in words:
        if len(current_line) + len(word) <= max_length:
            current_line += word + " "
        else:
            lines.append(current_line.strip())
            current_line = word + " "
    lines.append(current_line.strip())
    result = "\n".join(lines)
    return result
\end{codpy}
\end{framed}


\section{The \cpy{NSuniverse} class}
The universe set is the fundamental notion on which the definition of a neutrosophic set is founded on.
We have chosen to represent it by means of a list of strings.
\\
The corresponding class which implements such a notion is shortly described in the following UML class diagram.
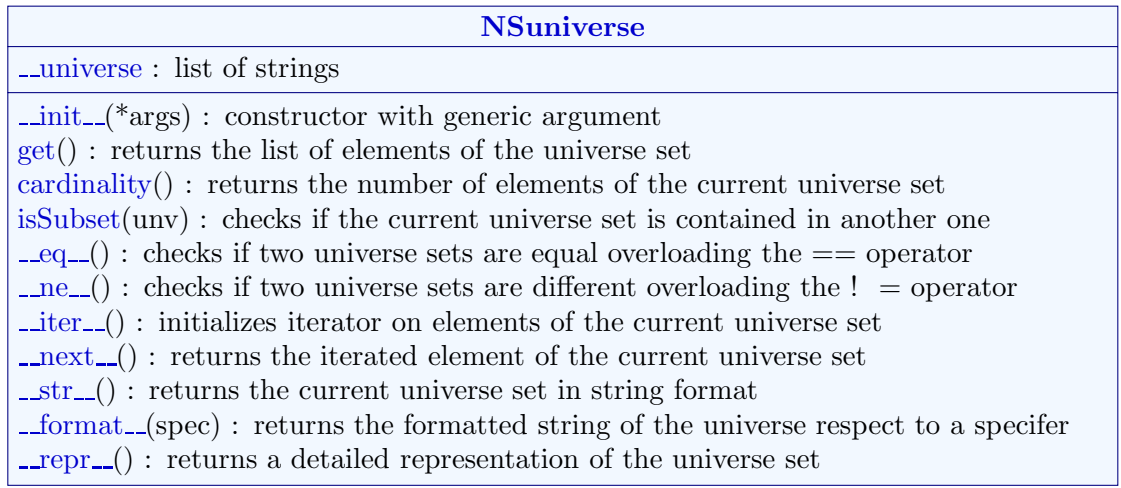
\vspace{-3mm}
\begin{figure}[H]
\begin{center}
\begin{tikzpicture}
  \begin{class}[text width=14.5cm]{NSuniverse}{0,0}
    \attribute{{\color{blu}\dunder universe} : list of strings}
    \operation{{\color{blu}\dunder init\dunder}(*args) : constructor with generic argument}
    \operation{{\color{blu}get}() : returns the list of elements of the universe set}
    \operation{{\color{blu}cardinality}() : returns the number of elements of the current universe set}
    \operation{{\color{blu}isSubset}(unv) : checks if the current universe set is contained in another one}
    \operation{{\color{blu}\dunder eq\dunder}() : checks if two universe sets are equal overloading the $==$ operator}
    \operation{{\color{blu}\dunder ne\dunder}() : checks if two universe sets are different overloading the $!=$ operator}
    \operation{{\color{blu}\dunder iter\dunder}() : initializes iterator on elements of the current universe set}
    \operation{{\color{blu}\dunder next\dunder}() : returns the iterated element of the current universe set}
    \operation{{\color{blu}\dunder str\dunder}() : returns the current universe set in string format}
    \operation{{\color{blu}\dunder format\dunder}(spec) : returns the formatted string of the universe respect to a specifer}
    \operation{{\color{blu}\dunder repr\dunder}() : returns a detailed representation of the universe set}
 \end{class}
\end{tikzpicture}
\end{center}
\label{dia:nsuniverse}
\end{figure}
\vspace{-4mm}

In order to ensure maximum usability and versatility in the use of this class,
the constructor method accepts string, lists, tuples, lists of elements of any length or another object \cpy{NSuniverse}
and proceeds to transform them into strings and store them in a list.
The basic steps of this method, expressed in pseudo-code, are described in the following algorithm.

\begin{algoritmo}
\caption{Constructor method of the class \cpy{NSuniverse}}
\SetKwFunction{MyFunction}{$\dunder$init$\dunder$}
\SetKwProg{Fn}{Function}{:}{}
\Fn{\MyFunction{args}}{
    Get the \textit{length} of \textit{args}\;
    \If{\textit{length} $=0$}{
        Raise an Exception\;
    }
    \ElseIf{\textit{length} $= 1$}{
        \If{\textit{args} \textbf{is} a list, a tuple or an object of the class}{
        Converts \textit{args} appropriately and stores it in \textit{universe}\;
        }
        \ElseIf{\textit{args} \textbf{is} a string}{
        Removes parentheses, commas and semicolons from \textit{args}, splits and gets a list of strings
        to store in \textit{universe}\;
        }
        \ElseIf{\textit{args} \textbf{is} a set}{
        Raise an Exception\;
        }
        \Else{
        Converts \textit{args} to string and creates a list with only this element to be stored in \textit{universe}\;
        }
    }
    \Else{
    Converts \textit{args} to a list of strings to be stored in \textit{universe}\;
}
    \If{\textit{universe} has repeated elements}{
        Raise an Exception\;
    }
   Stores \textit{universe} in the property $\dunder$\textit{universe}
}
\end{algoritmo}
\vspace{4mm}

The corresponding Python code of the constructor method of the \cpy{NSuniverse} class is given below.

\begin{framed}
\begin{codpy}
from .ns_util import NSreplace

class NSuniverse:

    def $\dunder$init$\dunder$(self, *args):
        universe = list()
        length = len(args)
        if length == 0:
            raise IndexError("the universe set must contain at least an element")
        elif length == 1:
            elem = args[0]
            if type(elem) in [list, tuple]:
                universe = [str(e) for e in elem]
            elif type(elem) == NSuniverse:
                universe = elem.get()
            elif type(elem) == str:
                sostituz = { "{":"", "}":"", "[":"", "]":"", "(":"", ")":"",
                             ",":" ", ";":" " }
                universe = NSreplace(elem, sostituz).split()
            elif type(elem) == set:
                raise ValueError("type set is not suitable because the elements of the universe set must be assigned in a specific order")
            else:
                universe = [str(elem)]
        else:
            for i in range(length):
                universe.append(str(args[i]))
        univset = set(universe)
        if len(universe) != len(univset):
            raise ValueError("the universe set cannot contain repeated elements")
        self.$\dunder$universe = universe
\end{codpy}
\end{framed}
\vspace{-3mm}


The constructor method also intercepts potential error situations in the definition
of universe sets such as attempting to define an empty set by calling it without any parameters
or that of inserting repeated elements (which conflicts with the usual set definition)
and in each of these cases raises an appropriate exception.


Let us observe that the flexibliity of the constructor method allows us to define a universe set
using various formats such as lists, tuples, strings, or simple enumerations of elements
without worrying about maintaining a rigid or uniform notation, which is particularly useful
to facilitate usability in interactive use.
For example, the universe set $\U = \{ 1,2,3,4,5 \}$ can be defined
as an object of the class \cpy{NSuniverse}
in any of the following ways mutually equivalent:
\begin{itemize}
\item \cpy{U=NSuniverse([1,2,3,4,5])} as a list,
\item \cpy{U=NSuniverse((1,2,3,4,5))} as a tuple,
\item \cpy{U=NSuniverse("1,2,3,4,5")} as a string of elements comma separated,
\item \cpy{U=NSuniverse("1;2;3;4;5")} as a string of elements separated by semicolon,
\item \cpy{U=NSuniverse("1 2 3 4 5")} as a string of elements separated by spaces,
\item \cpy{U=NSuniverse("1,2 3 4;5")} as a string of elements separated in various ways,
\item \cpy{U=NSuniverse("\{1,2,3,4,5\}")} as a string representing a set,
\item \cpy{U=NSuniverse("[1,2,3,4,5]")} as a string representing a list,
\item \cpy{U=NSuniverse("(1,2,3,4,5)")} as a string representing a tuple,
\item \cpy{U=NSuniverse("(1;2;3;4;5)")} using semicolon as separator,
\item \cpy{U=NSuniverse(1,2,3,4,5)} as a listing of numerical values only,
\item \cpy{U=NSuniverse("1",2,"3",4,"5")} in a mixed form,
\end{itemize}
as well as in different combinations of them.
\\
However, it is not allowed to define a universe set
by means of the \cpy{set} type of the Python language
(i.e., expressions such as \cpy{NSuniverse(\{1,2,3,4,5\})} are not accepted)
since it is an unordered data collection
and for a precise design choice the elements must be listed in a specific order,
feature this will prove valuable in simplifying and making consistent definitions
of both neutrosophic sets and mappings between them.

The class \cpy{NSuniverse} is equipped with very few basic methods, that is
\cpy{get()} which returns the list of strings corresponding to the instance of the universe set
exactly as it is stored internally in the class
and \cpy{cardinality()} which returns the number of elements present in the object
instantiated by the class.
\vspace{-3mm}
\begin{framed}
\begin{codpy}
    def get(self):
        return self.$\dunder$universe

    def cardinality(self):
        return len(self.$\dunder$universe)
\end{codpy}
\end{framed}
\vspace{-3mm}


The method \cpy{isSubset()} checks whether the current universe set
is contained in a second universe set passed as parameter
and returns the Boolean value \cpy{True} in the positive case.
\vspace{-3mm}
\begin{framed}
\begin{codpy}
    def isSubset(self, unv):
        setself = set(self.get())
        setunv = set(unv.get())
        result = setself.issubset(setunv)
        return result
\end{codpy}
\end{framed}
\vspace{-3mm}


To facilitate the comparison of two objects of type \cpy{NSuniverse},
the equality \cpy{==} and diversity \cpy{\!=} operators have been overloaded
using their corresponding special methods.
\vspace{-3mm}
\begin{framed}
\begin{codpy}
    def $\dunder$eq$\dunder$(self, unv):
        equal = (self.get() == unv.get())
        return equal

    def $\dunder$ne$\dunder$(self, unv):
        different = not (self == unv)
        return different
\end{codpy}
\end{framed}
\vspace{-3mm}


In order to be able to easily print on the screen objects of type \cpy{NSuniverse}
in text format and to provide a complete representation of them, the special methods
\cpy{$\dunder$str$\dunder$()} and
\linebreak 
\cpy{$\dunder$repr$\dunder$()} were defined as follows by using the overloading.
\vspace{-3mm}
\begin{framed}
\begin{codpy}
    def $\dunder$str$\dunder$(self):
        list_string_elements = [str(e) for e in self.$\dunder$universe]
        s = "{ " + ", ".join(list_string_elements) + " }"
        return s

    def $\dunder$repr$\dunder$(self):
        return f"Universe set: {str(self)}"
\end{codpy}
\end{framed}
\vspace{-3mm}


As an example, we show how the methods and operators described above
can be used not only in a client code, but also in the interactive mode
by means of the Python console:
\begin{codpyconsole}
>>> from pyns.ns_universe import NSuniverse
>>> U = NSuniverse("1", 2, 3, "4")
>>> print(U)
{ 1, 2, 3, 4 }
>>> print( U != NSuniverse([1,3,5]))
True
>>> V = NSuniverse(" ( a b c , d ; e )")
>>> print(V.cardinality())
5
>>> print(V.get())
['a', 'b', 'c', 'd', 'e']
>>> print(V)
{ a, b, c, d, e }
\end{codpyconsole}


Since in the following we will also need to make formatted prints
of objects of the universe set type according to a certain format specifier,
it is also necessary to redefine by overloading the special method
\cpy{$\dunder$format$\dunder$()}.

\begin{framed}
\begin{codpy}
    def $\dunder$format$\dunder$(self, spec):
        unvstr = str(self)
        result = f"{unvstr:{spec}}"
        return result
\end{codpy}
\end{framed}


Finally, to simplify the code of other classes devoted to neutrosophic sets and mappings,
it is useful to establish an iterator over the objects of the \cpy{NSuniverse} class.
This iterator should sequentially provide all and only the elements within a given universe set.
This is achieved by introducing a new property, $self.$\dunder$i$, to serve as the internal index
for the current element and redefining by overloading the special methods
\cpy{$\dunder$iter$\dunder$()} and \cpy{$\dunder$next$\dunder$()}
which are respectively intended to initialize the index of the iterator
and yield the element associated with the current index.

\begin{framed}
\begin{codpy}
    def $\dunder$iter$\dunder$(self):
        self.$\dunder$i = 0
        return self

    def $\dunder$next$\dunder$(self):
        if self.$\dunder$i < len(self.$\dunder$universe):
            elem = self.$\dunder$universe[self.$\dunder$i]
            self.$\dunder$i +=1
            return elem
        raise StopIteration
\end{codpy}
\end{framed}


Thanks to the introduction of the iterator on the class \cpy{NSuniverse} it will be, for example, possible to
handle loops directly on objects of type universe set, exactly as
happens with other standard Python types such as lists and tuples.
This approach will contribute to make the syntax of our code leaner and more understandable,
as highlighted in the following example.

\begin{framed}
\begin{codpy}
from pyns.ns_universe import NSuniverse

U = NSuniverse(" ( a b c , d ; e )")
for i, u in enumerate(U):
    print(f"- the {i}-th element is {u}")
\end{codpy}
\end{framed}

which produces output of the type:
\begin{codpyconsole}
- the 0-th element is a
- the 1-th element is b
- the 2-th element is c
- the 3-th element is d
- the 4-th element is e
\end{codpyconsole}


\section{The \cpy{NSset} class}

The representation of neutrosophic sets on a given universe set
is done by means of the \cpy{NSset} class which, obviously, uses the \cpy{NSuniverse} class.

The original definition of neutrosophic set, given in 1999 by Smarandache \cite{smarandache},
refers to the interval $]0^{-},1^{+}[$ of the nonstandard real numbers
and although it is consistent from a philosophical point of view,
unfortunately, it is not suitable to be used for approaching real-world problems.
For such a reason, in 2010, the same author, jointly with Wang, Zhang and Sunderraman \cite{wang},
also introduced the notion of single valued neutrosophic set which,
referring instead to the unit interval $[0,1]$ of the usual set of real numbers $\RR$,
can be usefully used in scientific and engineering applications.
In the following we will refer exclusively to single valued neutrosophic sets.

\begin{definition}{\rm\cite{wang}}
\label{def:singlevaluedneutrosophicset}
Let $\U$ be an universe set and $A\subseteq \U$,
a \df{single valued neutrosophic set} over $\U$ (\df{\nameSVNS} for short),
denoted by $\nNS{A}$, is a set of the form:
$$\ns{A} = \NSEXT{A}$$
where $\M{A} : \U \to I$, $\I{A} : \U \to I$ and $\NM{A} : \U \to I$
are the \df{membership function}, the \df{indeterminacy function} and the \df{non-membership function} of $A$,
respectively and $I=[0,1]$ be the unit interval of the real numbers.
For every $u \in \U$, $\DM{A}$, $\DI{A}$ and $\DNM{A}$ are said
the \df{degree of membership}, the \df{degree of indeterminacy} and the \df{degree of non-membership}
of $u$, respectively.
\end{definition}
\begin{definition}{\rm\cite{smarandache,wang}}
\label{def:neutrosophicsubset}
Let $\nNS{A}$ and $\nNS{B}$ be two \nameSVNS[s] over the universe set $\U$,
we say that $\ns{A}$ is a \df{neutrosophic subset} (or simply a subset) of $\ns{B}$
and we write $\ns{A} \NSsubseteq \ns{B}$
if, for every $u \in \U$, it results
$\DM{A} \le \DM{B}$,
$\DI{A} \le \DI{B}$
and $\DNM{A} \ge \DNM{B}$.
We also say that $\ns{A}$ is contained in $\ns{B}$ or that $\ns{B}$ contains $\ns{A}$
and we write $\ns{B} \NSsupseteq \ns{A}$ to denote that $\ns{B}$ is a \df{neutrosophic superset} of $\ns{A}$.
\end{definition}

\begin{definition}{\rm\cite{smarandache,wang}}
\label{def:neutrosophicequality}
Let $\nNS{A}$ and $\nNS{B}$ be two \nameSVNS[s] over the universe set $\U$.
We say that $\ns{A}$ is a \df{neutrosophically equal} (or simply equal) to $\ns{B}$ and we write
$\ns{A} \NSeq \ns{B}$
if $\ns{A} \NSsubseteq \ns{B}$ and $\ns{B} \NSsubseteq \ns{A}$.
\end{definition}

\begin{notation}
Let $\U$ be a set, $I=[0,1]$ the unit interval of the real numbers, for every $r \in I$,
with $\und{r}$ we denote the constant mapping $\und{r}: \U \to I$
defined by $\und{r}(u) = r$, for every $u \in \U$.
\end{notation}

\begin{definition}{\rm\cite{wang}}
\label{def:neutrosophicemptyset}
The \nameSVNS $\NSbase{\und{0}}{\und{0}}{\und{1}}$
is said to be the \df{neutrosophic empty set} over $\U$
and it is denoted by $\NSemptyset$,
or more precisely by $\NSemptyset[\U]$ in case it is necessary
to specify the corresponding universe set.
\end{definition}

\begin{definition}{\rm\cite{wang}}
\label{def:neutrosophicabsoluteset}
The \nameSVNS $\NSbase{\und{1}}{\und{1}}{\und{0}}$
is said to be the \df{neutrosophic absolute set} over $\U$
and it is denoted by $\NSabsoluteset$.
\end{definition}


In our class, the data structure used to represent a \nameSVNS is a dictionary
(also called associative array) which uses the elements of the universe set
as keys and associates them with a list of three floating-point numbers
corresponding to the degrees of membership, indeterminacy and non-membership respectively. 
This dictionary, referred as the \cpy{$\dunder$neutrosophicset} property,
is stored in conjunction with the universe set, referred as the \cpy{$\dunder$universe} property,
to which it is inseparably linked.
Indeed, it is no coincidence that the class \cpy{NSuniverse} does not provide
any method for allowing modifications (such as insertions or removals)
of the elements of an object of type universe set
since such operations could disrupt the consistency of the \nameSVNS[s] defined on it.
The \cpy{NSset} class is described by the following UML diagram.
\vspace{-4mm}

\begin{figure}[H]
\begin{center}
\begin{tikzpicture}
  \begin{class}[text width=14.5cm]{NSset}{0,0}
    \attribute{{\color{blu}\dunder universe} : object of the class NSuniverse}
    \attribute{{\color{blu}\dunder neutrosophicset} : dictionary of lists}
    \operation{{\color{blu}\dunder init\dunder}(*args) : constructor with generic arguments}
    \operation{{\color{blu}\dunder setDegree}(u,i,r) : assigns r to the i-th degree of element u}
    \operation{{\color{blu}setMembership}(u, mu) : assigns the membership degree to u}
    \operation{{\color{blu}setIndeterminacy}(u, sigma) : assigns the indeterminacy degree to u}
    \operation{{\color{blu}setNonMembership}(u, omega) : assigns the non-membership degree to u}
    \operation{{\color{blu}setElement}(u, triple) : assigns simultaneously the three degrees to u}
    \operation{{\color{blu}getUniverse}() : returns the universe of the \nameSVNS}
    \operation{{\color{blu}get}() : returns the dictionary containing the degrees of each element}
    \operation{{\color{blu}getElement}(u) : returns the triple the degrees of the element u}
    \operation{{\color{blu}\dunder getDegree}(u,i) : returns the i-th degree of element u}
    \operation{{\color{blu}getMembership}(u) : returns the membership degree of u}
    \operation{{\color{blu}getIndeterminacy}(u) : returns the indeterminacy degree of u}
    \operation{{\color{blu}getNonMembership}(u) : returns the non-membership degree of u}
    \operation{{\color{blu}setEmpty}() : makes equal to the neutrosophic empty set}
    \operation{{\color{blu}setAbsolute}() : makes equal to the absolute \nameSVNS}
    \operation{{\color{blu}cardinality}() : returns the number of elements of the \nameSVNS}
    \operation{{\color{blu}isNSsubset}(nset) : checks if is a neutrosophic subset of another one}
    \operation{{\color{blu}isNSsuperset}(nset) : checks if is a neutrosophic superset of another one}
    \operation{{\color{blu}NSunion}(nset) : returns the neutrosophic union with another one}
    \operation{{\color{blu}NSintersection}(nset) : returns the neutrosophic intersection with another one}
    \operation{{\color{blu}isNSdisjoint}(nset) : checks if is neutrosophically disjoint with another one}
    \operation{{\color{blu}NScomplement}() : returns the neutrosophic complement}
    \operation{{\color{blu}NSdifference}(nset) : returns the neutrosophic difference with another one}
    \operation{{\color{blu}\dunder eq\dunder}() : checks if two \nameSVNS[s] are equal overloading the $==$ operator}
    \operation{{\color{blu}\dunder ne\dunder}() : checks if two \nameSVNS[s] are different overloading the $!=$ operator}
    \operation{{\color{blu}\dunder add\dunder}(nset) : neutrosophic union overloading the $+$ operator}
    \operation{{\color{blu}\dunder and\dunder}(nset) : neutrosophic intersection overloading the $\&$ operator}
    \operation{{\color{blu}\dunder invert\dunder}() : neutrosophic complement overloading the $\sim$ operator}
    \operation{{\color{blu}\dunder sub\dunder}(nset) : neutrosophic difference overloading the $-$ operator}
    \operation{{\color{blu}\dunder le\dunder}(nset) : neutrosophic subset overloading the $<=$ operator}
    \operation{{\color{blu}\dunder ge\dunder}(nset) : neutrosophic superset overloading the $>=$ operator}
    \operation{{\color{blu}\dunder str\dunder}() : returns the \nameSVNS in string format}
    \operation{{\color{blu}\dunder format\dunder}(spec) : returns the formatted string of the \nameSVNS respect to a specifer}
    \operation{{\color{blu}\dunder repr\dunder}() : returns a detailed representation of the \nameSVNS}
 \end{class}
\end{tikzpicture}
\end{center}
\label{dia:nsset}
\end{figure}

The constructor method of this class accepts one or two parameters and allows us
to define a \nameSVNS in several different ways:
\begin{itemize}
\item in the form \cpy{NSset($\textit{\textrm{universe set}}$)}
assigning to it as its only parameter the universe set (expressed as a list, tuple, or string) over which it is defined
and thus creating an empty \nameSVNS $\NSemptyset$,
\item in the form \cpy{NSset($\textit{\textrm{universe set}},\textrm{\textit{values}}$)} passing two parameters,
the first of which is a universe set and the second
an enumeration (expressed as a list, tuple, or string) of triples of real values
representing the degree of membership, indeterminacy and non-membership of all the elements of the universe set, or
\item in the form \cpy{NSset($\textit{\textrm{neutrosophic set}}$)} by copying another object of the type \cpy{NSset}.
\end{itemize}
\vfill

The basic steps of this method, expressed in pseudo-code, are described in the following algorithm.
\vspace{3mm}

\begin{algoritmo}
\caption{Constructor method of the class \cpy{NSset}}
\SetKwFunction{MyFunction}{$\dunder$init$\dunder$}
\SetKwProg{Fn}{Function}{:}{}
\Fn{\MyFunction{args}}{
    Create a dictionary \textit{neutrosophicset}\;
    Get the \textit{length} of \textit{args}\;
    \If{\textit{length} $=1$}{
       \If{\textit{args} \textbf{is} a list, a tuple, a string or an object of the class \cpy{NSuniverse}}{
          Create \textit{universe} from \textit{args} and set \textit{neutrosophicset} empty\;
       }
       \ElseIf{\textit{args} \textbf{is} an object of the class \cpy{NSset}}{
          Get \textit{universe} and  \textit{neutrosophicset} from \textit{args}\;
       }
       \Else{
          Raise an Exception\;
       }
    }
    \ElseIf{\textit{length} $=2$}{
       Use the same constructor with the first parameter of \textit{args} to obtain an object
       of type \cpy{NSset} from which to derive \textit{universe}
       and set the list \textit{values} equal to the second parameter of \textit{args}\;
       \If{\textit{values} \textbf{is} a list or a tuple}{
          \If{length of \textit{values} is different from the length of \textit{universe}}{
          Raise an Exception\;
          }
          Assigns to each element of \textit{universe} the values of the corresponding triple of \textit{values}\;
       }
       \ElseIf{\textit{args} \textbf{is} a string}{
          Converts \textit{args} to a list of triples and uses the same constructor
          with \textit{universe} and such a list
          to obtain an object of type \cpy{NSset} from which to take \textit{neutrosophicset}\;
       }
       \Else{
          Raise an Exception\;
       }
    }
    \Else{
        Raise an Exception\;
    }
   Stores \textit{universe} and  \textit{neutrosophicset} in the properties
   $\dunder$\textit{universe} and $\dunder$\textit{neutrosophicset} respectively\;
}
\end{algoritmo}
\vspace{4mm}


Note how every possible error condition
-- such as, for example, attempting to pass a variable of type other than \cpy{NSuniverse}
or a list of values not consisting of triples of real numbers included in the $[0,1]$ interval
or, again, using it with fewer than one or more than two parameters --
is intercepted in the code and reported to the client
by raising an appropriate exception.


The Python code corresponding to the constructor method of the \cpy{NS_set} class is given below.
\vspace{-3mm}

\begin{framed}
\begin{codpy}
from .ns_universe import NSuniverse
from .ns_util import NSreplace, NSstringtoTriplesList, NSsplitText

class NSset:

    degreename = ["membership", "indeterminacy", "non-membership"]
    reprmaxlength = 64

    def $\dunder$init$\dunder$(self, *args):
        neutrosophicset = dict()
        length = len(args)
        if length == 1:
            element = args[0]
            if type(element) in [list, tuple, str, NSuniverse]:
                universe = NSuniverse(element)
                for e in universe.get():
                    neutrosophicset[e] = [0,0,1]
            elif type(element) == NSset:
                universe = element.getUniverse()
                for e in universe:
                    neutrosophicset[e] = element.getElement(e)
            else:
                raise ValueError("value not compatible with the type universe set")
        elif length == 2:
            nset = NSset(args[0])
            universe = nset.getUniverse()
            values = args[1]
            if type(values) in [list ,tuple]:
                if len(values) != len(universe):
                    raise IndexError("the number of value triples does not correspond with the number of elements")
                for i in range(len(universe)):
                    elem = universe[i]
                    t = values[i]
                    if type(t) not in [tuple,list] or len(t) !=3:
                        raise IndexError("the second parameter of the constructor method must contain only triple")
                    t = [float(t[j]) for j in range(3)]
                    for j in range(3):
                        if not 0 <= t[j] <= 1:
                            raise ValueError(f"incompatible {self.degreename[j]} degree value")
                    neutrosophicset[elem] = t
            elif type(values) == str:
                tpl_list = NSstringtoTriplesList(values)
                nset = NSset(universe, tpl_list)
                neutrosophicset = nset.get()
            else:
                raise ValueError("the second parameter of the constructor method must contain a list of triples of real numbers")
        else:
            raise IndexError("the number of parameters do not match those of the constructor method")
        self.$\dunder$universe = NSuniverse(universe)
        self.$\dunder$neutrosophicset = neutrosophicset
\end{codpy}
\end{framed}

Let us observe that to enable the parsing of strings containing triples with the values of membership degrees, indeterminacy
and non-membership, variously expressed, were used the function \cpy{NSstringToTriplesList}
contained in the utility file \cpy{NS_util.py}.

Similar to what we have seen previously for objects of type universe set,
the constructor method of the class \cpy{NSset} also allows for a multiplicity of expressions
thanks to which we can define neutrosophic sets with a direct and informal notation
by means of lists, tuples or strings of elements separated indifferently by commas or semicolons.
\\
For example, the \nameSVNS $\NS{A}$ over the universe set $\U = \{ a,b,c \}$
defined by:
$$\ANG{\frac{a}{(0.5,0.3,0.2)}, \frac{b}{(0.6,0.2,0.3)}, \frac{c}{(0.4,0.2,0.7)} }$$
can be defined as an object of the class \cpy{NSset}
in any of the following ways mutually equivalent:
\begin{itemize}
\item \cpy{A=NSset("a,b,c",[(0.5,0.3,0.2),(0.6,0.2,0.3),(0.4,0.2,0.7)]) } as a list of tuples,
\item \cpy{A=NSset("a,b,c",[[0.5,0.3,0.2],[0.6,0.2,0.3],[0.4,0.2,0.7]]) } as a list of lists,
\item \cpy{A=NSset("a,b,c",[[0.5,0.3,0.2],(0.6,0.2,0.3),[0.4,0.2,0.7]]) } as a mixed list of lists and tuples,
\item \cpy{A=NSset("a,b,c",([0.5,0.3,0.2],[0.6,0.2,0.3],[0.4,0.2,0.7])) } as a tuple of lists,
\item \cpy{A=NSset("a,b,c",((0.5,0.3,0.2),(0.6,0.2,0.3),(0.4,0.2,0.7))) } as a tuple of tuples,
\item \cpy{A=NSset("a,b,c",((0.5,0.3,0.2),[0.6,0.2,0.3],(0.4,0.2,0.7))) } as a mixed tuple of tuples and lists,
\item \cpy{A=NSset("a,b,c","[0.5,0.3,0.2],(0.6,0.2,0.3);[0.4,0.2,0.7]") } as a string containing lists and tuples
\end{itemize}
where the universe set can also be expressed in any equivalent form as
\cpy{U=NSuniverse("\{a,b,c\}")},
\cpy{U=NSuniverse("[a,b;c]")}
or \cpy{U=NSuniverse("(a;b,c)")}
so that it can be used later in the definition of the \nameSVNS in the form:
\begin{itemize}
\item \cpy{A=NSset(U,"[0.5,0.3,0.2],(0.6,0.2,0.3);[0.4,0.2,0.7]") }
\end{itemize}


A \nameSVNS $\NS{A}$ already defined over a universe set $\U$ can be subsequently modified
by assigning to each of its generic elements $u \in \U$ its degrees of membership $\textit{mu} = \DM[u]{A}$,
indeterminacy $\textit{sigma} = \DI[u]{A}$ and non-membership $\textit{omega} = \DNM[u]{A})$
separately through the methods:
\begin{itemize}
\item \cpy{setMembership(self,$\textit{\textrm{u}}$,$\textit{\textrm{mu}}$)}
\item \cpy{setIndeterminacy(self,$\textit{\textrm{u}}$,$\textit{\textrm{sigma}}$)}, and
\item \cpy{setNonMembership(self,$\textit{\textrm{u}}$,$\textit{\textrm{omega}}$)}
\end{itemize}
or by assigning in one shot the entire $\textit{triple}=\left( \DM[u]{A}, \DI[u]{A}, \DNM[u]{A} \right)$
of values using the method:
\begin{itemize}
\item \cpy{setElement(self,$\textit{\textrm{u}}$,$\textit{\textrm{triple}}$)}
\end{itemize}
all of which are based on the private method
\cpy{$\dunder$setDegree(self,$\textit{\textrm{u}}$,$\textit{\textrm{i}}$,$\textit{\textrm{v}}$)}
that assigns the value $v$ to the $i$-th degree (for $i=0,1,2$, which correspond in the order to
membership, indeterminacy and non-membership degree) of a given element $u$
of the current \nameSVNS.
Obviously, in the latter method we take into account the fact that the element $u$ must
belong to the corresponding universe set and that the membership degrees must be real values included
in the unit interval $I$ and if not, appropriate exceptions will be raised.
\vspace{-3mm}

\begin{framed}
\begin{codpy}
    def $\dunder$setDegree(self, u, i, r):
        u = str(u)
        if u not in self.getUniverse():
            raise IndexError('non-existent element')
        r = float(r)
        if not (0 <= r <= 1):
            raise ValueError(f"incompatible {self.degreename[i]} degree value")
        self.$\dunder$neutrosophicset[u][i] = r

    def setMembership(self, u, mu):
        self.$\dunder$setDegree(u, 0, mu)

    def setIndeterminacy(self, u, sigma):
        self.$\dunder$setDegree(u, 1, sigma)

    def setNonMembership(self, u, omega):
        self.$\dunder$setDegree(u, 2, omega)

    def setElement(self, u, triple):
        if type(triple) == str:
            sostituz = { "(":"", ")":"", ",":" ", ";":" " }
            triple = NSreplace(triple, sostituz).split()
        else:
            triple = list(triple)
        if len(triple) != 3:
            raise ValueError('error in the number of parameters passed')
        triple = [float(e) for e in triple]
        for i in range(3):
            self.$\dunder$setDegree(u, i, triple[i])
\end{codpy}
\end{framed}


Three basic methods called \cpy{getUniverse()}, \cpy{get()} and \cpy{getElement($\textit{\textrm{u}}$)}
respectively return to us the universe set of a given \nameSVNS
as a string list of its elements, the neutrosophic set itself as a dictionary
having for keys the elements of the universe and for values the triples of the degrees
of membership, indeterminacy and non-membership as well as the triple of the degrees of a given
element $u \in \U$.
\vspace{-3mm}

\begin{framed}
\begin{codpy}
    def getUniverse(self):
        return self.$\dunder$universe.get()

    def get(self):
        return self.$\dunder$neutrosophicset

    def getElement(self, u):
        u = str(u)
        if u not in self.getUniverse():
            raise IndexError('non-existent element')
        return self.$\dunder$neutrosophicset[u]
\end{codpy}
\end{framed}
\vspace{-3mm}


Given a \nameSVNS $\NS{A}$ over a universe set $\U$,
The values of the degrees of membership $\DM[u]{A}$, indeterminacy $\DI[u]{A}$
and non-membership $\DNM[u]{A}$ of a generic element $u \in \U$ can
be obtained by the methods:
\begin{itemize}
\item \cpy{getMembership($\textit{\textrm{u}}$)},
\item \cpy{getIndeterminacy($\textit{\textrm{u}}$)}, and
\item \cpy{getNonMembership($\textit{\textrm{u}}$)}
\end{itemize}
which are all based on the private method
\cpy{$\dunder$getDegree(self,$\textit{\textrm{u}}$,$\textit{\textrm{i}}$)}
that returns the value of the $i$-th degree (for $i=0,1,2$, which correspond in the order to
membership, indeterminacy and non-membership degree) of a given element $u$
of the current \nameSVNS.
Let us note that in the latter method, we consider the prerequisite that the element $u$
must belong to the corresponding universe set and if this condition is not satisfied
a suitable exception will be raised.
\vspace{-3mm}

\begin{framed}
\begin{codpy}
    def $\dunder$getDegree(self, u, i):
        u = str(u)
        if u not in self.getUniverse():
            raise IndexError('non-existent element')
        return self.$\dunder$neutrosophicset[u][i]

    def getMembership(self, u):
        return self.$\dunder$getDegree(u, 0)

    def getIndeterminacy(self, u):
        return self.$\dunder$getDegree(u, 1)

    def getNonMembership(self, u):
        return self.$\dunder$getDegree(u, 2)
\end{codpy}
\end{framed}

\vspace{-3mm}


In case we need the empty neutrosophic set $\NSemptyset$
or the neutrosophic absolute set $\NSabsoluteset$ over an universe set $\U$
we can refer to the methods \cpy{setEmpty()} and \cpy{setAbsolute()}, respectively.
\vspace{-9mm}

\begin{framed}
\begin{codpy}
    def setEmpty(self):
        for e in self.$\dunder$universe.get():
            self.$\dunder$neutrosophicset[e] = [0, 0, 1]

    def setAbsolute(self):
        for e in self.$\dunder$universe.get():
            self.$\dunder$neutrosophicset[e] = [1, 1, 0]
\end{codpy}
\end{framed}


The on-screen printing in text format of objects of type \cpy{NSset}
as well as their complete representation is achieved by overloading the special methods
\cpy{$\dunder$str$\dunder$} and \cpy{$\dunder$repr$\dunder$}
as follows.
\vspace{-9mm}

\begin{framed}
\begin{codpy}
    def $\dunder$str$\dunder$(self, tabularFormat=False):
        if tabularFormat == True:
            (dashes, elemwidth, valwidth) = ("-"*64, 10, 14)
            s = "\n            |   membership   |  indeterminacy | non-membership |\n" + dashes + "\n"
            for e in self.getUniverse():
                (mu, sigma, omega) = self.getElement(e)
                s += f" {str(e):{elemwidth}} | {mu:{valwidth}} | {sigma:{valwidth}} | {omega:{valwidth}} |\n"
            s += dashes + "\n"
        else:
            elems = []
            for e in self.getUniverse():
                (mu, sigma, omega) = self.getElement(e)
                elems.append(f"{e}/({mu},{sigma},{omega})")
            s = "< " + ", ".join(elems) + " >"
            s = NSsplitText(s, self.reprmaxlength)
        return s

    def $\dunder$repr$\dunder$(self):
        return f"Neutrosophic set: {str(self)}"
\end{codpy}
\end{framed}
\vspace{-3mm}

In particular, the special method \cpy{$\dunder$str$\dunder$}
allow us to print on the screen a \nameSVNS in both the simplified representation
(which is the default option) and in the clearer and more extensive tabular representation.

Furthermore, in order to be able to choose to print a \nameSVNS in the simplified representation
or in the tabular one even in interactive use or writing client code,
it was chosen to redefine the special method \cpy{$\dunder$format$\dunder$}
so that it recognizes the new custom format specifier
\cpy{t} that corresponds to printing in tabular format objects of type \cpy{NSset}.
\vspace{-3mm}

\begin{framed}
\begin{codpy}
    def $\dunder$format$\dunder$(self, spec):
        if spec == "t":
            result = self.$\dunder$str$\dunder$(tabularFormat=True)
        else:
            result = self.$\dunder$str$\dunder$(tabularFormat=False)
        return result
\end{codpy}
\end{framed}
\vspace{-3mm}


Thanks to the redefinition by overloading of the special methods \cpy{$\dunder$eq$\dunder$}
and \cpy{$\dunder$ne$\dunder$} we can use the operators
of equality \cpy{==} and diversity \cpy{\!=} directly to objects of type \cpy{NSset}.
\vspace{-3mm}

\begin{framed}
\begin{codpy}
    def $\dunder$eq$\dunder$(self, nset):
        if self.getUniverse() != nset.getUniverse():
            raise ValueError("the two neutrosophic sets cannot be defined on different universe sets")
        equal = self.isNSsubset(nset) and nset.isNSsubset(self)
        return equal

    def $\dunder$ne$\dunder$(self, nset):
        if self.getUniverse() != nset.getUniverse():
            raise ValueError("the two neutrosophic sets cannot be defined on different universe sets")
        different = not (self == nset)
        return different
\end{codpy}
\end{framed}
\vspace{-2mm}


In order to better illustrate how the above methods are used,
let us consider the following example of code executed interactively in the Python console.

\begin{codpyconsole}
>>> from pyns.ns_universe import NSuniverse
>>> from pyns.ns_set import NSset
>>> U = NSuniverse("a,b,c")
>>> A = NSset(U)
>>> A.setElement('a', (0.8,0.2,0.1))
>>> A.setElement('c', (0.3,0.2,0.4))
>>> A.getMembership('a')
0.8
>>> A.getNonMembership('c')
0.4
>>> print(A.getElement(c))
[0.3, 0.2, 0.4]
>>> A.setIndeterminacy('b',0.9)
>>> print(A)
< a/(0.8,0.2,0.1), b/(0.0,0.9,1.0), c/(0.3,0.2,0.4) >
>>> print(f"{A:t}")
            |   membership   |  indeterminacy | non-membership |
----------------------------------------------------------------
 a          |            0.8 |            0.2 |            0.1 |
 b          |              0 |            0.9 |              1 |
 c          |            0.3 |            0.2 |            0.4 |
----------------------------------------------------------------
>>> A.setAbsolute()
>>> print(A)
< a/(1,1,0), b/(1,1,0), c/(1,1,0) >
\end{codpyconsole}
\vspace{-2mm}


From the example above, one might assume that creating universe sets
and \nameSVNS[s] requires manual definition.
However, the open structure of our framework actually enables us to define objects
of type \cpy{NSuniverse} and \cpy{NSset} dynamically within the code,
commonly referred to as defining them 'on the fly'.
This dynamic approach is especially advantageous when dealing with \nameSVNS[s]
of considerable cardinality, as illustrated in the following Python code.
\vspace{-3mm}

\begin{framed}
\begin{codpy}
from pyns.ns_universe import NSuniverse
from pyns.ns_set import NSset
from random import random

lst = [(i,j) for i in range(1,6) for j in range(1,4)]

U = NSuniverse(lst)
A = NSset(U)

for u in U.get():
    triple = [round(random(), 2) for k in range(3)]
    A.setElement(u, triple)

print(f"The following SVNS-set has cardinality {A.cardinality()}: {A:t}")
\end{codpy}
\end{framed}

\noindent
which produces output of the type:
\begin{codpyconsole}
The following SVNS-set has cardinality 15:
            |   membership   |  indeterminacy | non-membership |
----------------------------------------------------------------
 (1, 1)     |           0.55 |            0.1 |            0.5 |
 (1, 2)     |           0.39 |           0.92 |           0.09 |
 (1, 3)     |           0.33 |           0.29 |           0.25 |
 (2, 1)     |           0.16 |            0.9 |           0.43 |
 (2, 2)     |           0.71 |           0.52 |           0.33 |
 (2, 3)     |           0.65 |           0.38 |           0.04 |
 (3, 1)     |           0.95 |           0.14 |           0.94 |
 (3, 2)     |           0.74 |           0.02 |           0.01 |
 (3, 3)     |           0.77 |           0.63 |           0.19 |
 (4, 1)     |           0.18 |           0.75 |           0.15 |
 (4, 2)     |           0.49 |           0.92 |           0.75 |
 (4, 3)     |           0.34 |           0.17 |           0.88 |
 (5, 1)     |           0.88 |            0.6 |           0.83 |
 (5, 2)     |            0.5 |           0.56 |            0.8 |
 (5, 3)     |           0.38 |           0.47 |            0.2 |
----------------------------------------------------------------
\end{codpyconsole}


To verify that a \nameSVNS expressed as an object of type \cpy{NSset} is neutrosophically contained
in another \nameSVNS, we may resort to the method \cpy{isNSsubset($\textit{\textrm{nset}}$)} which,
similarly to the built-in \cpy{issubset()} method available for objects of type \cpy{set},
returns the Boolean value \cpy{True} if the current \nameSVNS is neutrosophically contained in the
second \nameSVNS \textit{nset} passed as a parameter or the value \cpy{False} otherwise.
\vspace{-3mm}

\begin{framed}
\begin{codpy}
    def isNSsubset(self, nset):
        if self.getUniverse() != nset.getUniverse():
            raise ValueError("the two neutrosophic sets cannot be defined on different universe sets")
        if self.getUniverse() != nset.getUniverse():
            return False
        else:
            result = True
            for e in self.getUniverse():
                (muA, sigmaA, omegaA) = self.getElement(e)
                (muB, sigmaB, omegaB) = nset.getElement(e)
                if (muA > muB) or (sigmaA > sigmaB) or (omegaA < omegaB):
                    result = False
                    break
            return result
\end{codpy}
\end{framed}

Based on \cpy{isNSsubset}, it is then immediate to define the method \cpy{isNSsuperset($\textit{\textrm{nset}}$)}
(analogous to the built-in \cpy{issuperset} method) which returns \cpy{True}
if the current \nameSVNS neutrosophically contains the
second \nameSVNS \textit{nset} passed as a parameter or \cpy{False} otherwise.
\vspace{-3mm}

\begin{framed}
\begin{codpy}
    def isNSsuperset(self, nset):
        if self.getUniverse() != nset.getUniverse():
            raise ValueError("the two neutrosophic sets cannot be defined on different universe sets")
        return nset.isNSsubset(self)
\end{codpy}
\end{framed}

In both cases, it is preliminarily verified that the two \nameSVNS[s] are
defined on the same universe set and if not an appropriate exception is raised.

The following code executed in interactive mode in the Python console
illustrates the use of the methods just described.

\begin{codpyconsole}
>>> from pyns.ns_universe import NSuniverse
>>> from pyns.ns_set import NSset
>>> A = NSset("a, b, c", "(0.3,0,0.5), (0.7,0.2,0.2), (0.1,0.5,0.4)")
>>> print(A)
< a/(0.3,0.0,0.5), b/(0.7,0.2,0.2), c/(0.1,0.5,0.4) >
>>> B = NSset("a, b, c", "(0.4,0.2,0.3), (0.8,0.3,0.1), (0.2,0.5,0.2)")
>>> print(B)
< a/(0.4,0.2,0.3), b/(0.8,0.3,0.1), c/(0.2,0.5,0.2) >
>>> print(A.isNSsubset(B))
True
>>> print(A.isNSsuperset(B))
False
\end{codpyconsole}


\begin{definition}{\rm\cite{salama2013}}
\label{def:neutrosophicunion}
The \df{neutrosophic union} of two \nameSVNS[s]
$\nNS{A}$ and $\nNS{B}$, denoted by $\ns{A} \NScup \ns{B}$, is the neutrosophic set defined by
$\NSbase{\M{A} \vee \M{B}}{\I{A} \vee \I{B}}{\NM{A} \wedge \NM{B}}$.
\end{definition}
\vspace{-3mm}

\begin{definition}{\rm\cite{salama2013}}
\label{def:neutrosophicintersection}
The \df{neutrosophic intersection} of two \nameSVNS[s]
$\nNS{A}$ and $\nNS{B}$, denoted by $\ns{A} \NScap \ns{B}$, is the neutrosophic set defined by
$\NSbase{\M{A} \wedge \M{B}}{\I{A} \wedge \I{B}}{\NM{A} \vee \NM{B}}$.
\end{definition}
\vspace{-3mm}

\begin{definition}{\rm\cite{wang}}
\label{def:neutrosophicdisjoint}
Let $\nNS{A}$ and $\nNS{B}$ be two \nameSVNS[s] over $\U$.
We say that $\ns{A}$ and $\ns{B}$ are \df{neutrosophically disjoint}
if $\ns{A} \NScap \ns{B} \NSeq \NSemptyset$.
On the contrary, if $\ns{A} \NScap \ns{B} \NSneq \NSemptyset$
we say that $\ns{A}$ \df{neutrosophically meets} $\ns{B}$
(or that $\ns{A}$ and $\ns{B}$ neutrosophically meet each other).
\end{definition}

Within our \cpy{NSset} class, the neutrosophic union and neutrosophic intersection
were implemented through the methods \cpy{NSunion()} and \cpy{NSintersection()}, respectively.
These methods mirror the built-in Python methods \cpy{union} e \cpy{intersection} for objects of type \cpy{set}
and they are both based on the private method
\cpy{$\dunder$NSoperation(self, $\textit{\textrm{nset}}$, $\textit{\textrm{fm}}$,
$\textit{\textrm{fs}}$, $\textit{\textrm{fo}}$)}
corresponding to a generic operation and that returns the neutrosophic set
obtained from the current \nameSVNS and the second \nameSVNS \textit{nset}
by applying the three functions \textit{fm}, \textit{fs}, \textit{fo} passed as parameters
to their membership, indeterminacy and non-membership degrees respectively.

\begin{framed}
\begin{codpy}
    def $\dunder$NSoperation(self, nset, fm, fs, fo):
        if self.getUniverse() != nset.getUniverse():
            raise ValueError("the two neutrosophic sets cannot be defined on different universe sets")
        if callable(fm) == False or callable(fs) == False or callable(fo) == False:
            raise  ValueError("the last three parameters must be functions")
        C = NSset(self.$\dunder$universe)
        for e in self.getUniverse():
            (muA, sigmaA, omegaA) = self.getElement(e)
            (muB, sigmaB, omegaB) = nset.getElement(e)
            triple = [fm(muA, muB), fs(sigmaA, sigmaB), fo(omegaA, omegaB)]
            C.setElement(e, triple)
        return C

    def NSunion(self, nset):
        C = self.$\dunder$NSoperation(nset, max, max, min)
        return C

    def NSintersection(self, nset):
        C = self.$\dunder$NSoperation(nset, min, min, max)
        return C
\end{codpy}
\end{framed}
\vspace{-2mm}


We illustrate the above methods with an example of code
executed interactively in the Python console.

\begin{codpyconsole}
>>> from pyns.ns_universe import NSuniverse
>>> from pyns.ns_set import NSset
>>> U = NSuniverse("a, b, c")
>>> A = NSset(U,"(0.3,0,0.5), (0.7,0.2,0.2), (0.1,0.5,0.4)")
>>> print(A)
< a/(0.3,0.0,0.5), b/(0.7,0.2,0.2), c/(0.1,0.5,0.4) >
>>> B = NSset(U,"(0.4,0.2,0.3), (0.8,0.3,0.1), (0.2,0.5,0.2)")
>>> print(B)
a/(0.4,0.2,0.3), b/(0.8,0.3,0.1), c/(0.2,0.5,0.2) >
>>> C = A.NSunion(B)
>>> print(C)
< a/(0.4,0.2,0.3), b/(0.8,0.3,0.1), c/(0.2,0.5,0.2) >
>>> D = A.NSintersection(B)
>>> print(D)
< a/(0.3,0.0,0.5), b/(0.7,0.2,0.2), c/(0.1,0.5,0.4) >
\end{codpyconsole}
\vspace{-1mm}


The method \cpy{isNSdisjoint(nset)} returns the Boolean value \cpy{True}
if the current \nameSVNS is neutrosophically disjoint from the second \nameSVNS passed as parameter
or the value \cpy{False} otherwise.
\vspace{-3mm}

\begin{framed}
\begin{codpy}
    def isNSdisjoint(self, nset):
        nsempty = NSset(self.$\dunder$universe)
        disjoint = self.NSintersection(nset) == nsempty
        return disjoint
\end{codpy}
\end{framed}

An example of the use of this method is provided in the following code executed in the console
Python interactive.

\begin{codpyconsole}
>>> from pyns.ns_universe import NSuniverse
>>> from pyns.ns_set import NSset
>>> A = NSset("a, b, c", "(0.3,0,0.5), (0.7,0,1), (0,0.5,1)")
>>> B = NSset("a, b, c", "(0,0.8,1), (0,0.1,0.2), (0.1,0.0,0.4)")
>>> print(A.isNSdisjoint(B))
True
\end{codpyconsole}


\begin{definition}{\rm\cite{smarandache,wang}}
\label{def:neutrosophiccomplement}
Let $\nNS{A}$ be a \nameSVNS over the universe set $\U$,
the \df{neutrosophic complement} (or, simply, the complement) of $\ns{A}$, denoted by $\ns{A}^\NScompl$,
is the \nameSVNS
$\ns{A}^\NScompl = \NSbase{\NM{A}}{\und{1}-\I{A}}{\M{A}}$
that is
$\ns{A}^\NScompl = \left\{ \left( u, \DNM{A}, 1-\DI{A}, \DM{A} \right) : u \in \U \right\}$.
\end{definition}

Although the neutrosophic difference of two \nameSVNS[s] $\ns{A}$ and $\ns{B}$ can be
defined (in analogy with ordinary sets)
as the neutrosophic intersection of the first set with the neutrosophic complement of the second set,
that is, as $\ns{A} \NScap \ns{B}^\NScompl$, for our purposes it is preferable to
provide an explicit and operational definition.
\vspace{-3mm}

\begin{definition}
\label{def:neutrosophicdifference}
The \df{neutrosophic difference} of two \nameSVNS[s]
$\nNS{A}$ and $\nNS{B}$, denoted by $\ns{A} \NSsetminus \ns{B}$, is the neutrosophic set defined by
$\NSbase{\M{A} \wedge \NM{B}}{\I{A} \wedge(\und{1}-\I{B})}{\NM{A} \vee \M{B}}$.
\end{definition}
\vspace{-3mm}

The latter operations have also been implemented in the class \cpy{NSset}
through the methods \cpy{NScomplement()} and \cpy{NSdifference()}.
\vspace{-3mm}

\begin{framed}
\begin{codpy}
    def NScomplement(self):
        C = NSset(self.$\dunder$universe)
        for e in self.getUniverse():
            (muA, sigmaA, omegaA) = self.getElement(e)
            triple = [omegaA, 1 - sigmaA, muA]
            C.setElement(e, triple)
        return C

    def NSdifference(self, nset):
        if self.getUniverse() != nset.getUniverse():
            raise ValueError("the two neutrosophic sets cannot be defined on different universe sets")
        C = NSset(self.$\dunder$universe)
        for e in self.getUniverse():
            (muA, sigmaA, omegaA) = self.getElement(e)
            (muB, sigmaB, omegaB) = nset.getElement(e)
            triple = [min(muA,omegaB), min(sigmaA,1-sigmaB), max(omegaA,muB)]
            C.setElement(e, triple)
        return C
\end{codpy}
\end{framed}

The following code example executed interactively in the Python console
illustrates the use of the two methods that have been just described.

\begin{codpyconsole}
>>> from pyns.ns_universe import NSuniverse
>>> from pyns.ns_set import NSset
>>> U = NSuniverse('a','b','c')
>>> A = NSset(U, "(0.5,0.3,0.2), (0.6,0.2,0.3), (0.4,0.2,0.7)")
>>> print(A)
< a/(0.5,0.3,0.2), b/(0.6,0.2,0.3), c/(0.4,0.2,0.7) >
>>> B = NSset(U, "(0.2,0.2,0.2), (0.4,0.1,0.6), (0.8,0.3,0.1)")
>>> print(B)
< a/(0.2,0.2,0.2), b/(0.4,0.1,0.6), c/(0.8,0.3,0.1) >
>>> C = A.NScomplement()
>>> print(C)
< a/(0.2,0.7,0.5), b/(0.3,0.8,0.6), c/(0.7,0.8,0.4) >
>>> D = A.NSdifference(B)
>>> print(D)
< a/(0.2,0.3,0.2), b/(0.6,0.2,0.4), c/(0.1,0.2,0.8) >
\end{codpyconsole}


To facilitate even more streamlined and intuitive use of neutrosophic set operations,
especially in the interactive use of the framework,
overloading was then used to redefine
the special methods
\cpy{$\dunder$add$\dunder$},
\cpysimple{$\dunder$and$\dunder$},
\cpy{$\dunder$invert$\dunder$},
\cpy{$\dunder$sub$\dunder$},
\cpy{$\dunder$le$\dunder$},
and
\cpy{$\dunder$ge$\dunder$},
respectively referred to the operators
\cpy{$+$}, \cpy{$\&$}, \cpy{$\sim$}, \cpy{$-$}, \cpy{$<=$}, and \cpy{$>=$}
for use with objects of type \cpy{NSset}
by making them coincide with the methods
\cpy{NSunion()}, \cpy{NSintersection()}, \cpy{NScomplement()},
\cpy{NSdifference()}, \cpy{isNSsubset()} and \cpy{isNSsuperset()},
thus obtaining the correspondence summarized in the following table.
\vspace{-3mm}

\begin{figure}[H]
\begin{center}
\begin{tabular}{|l|l|c|c|}
\hline
 & \qquad \textbf{class method} \qquad & \textbf{symbol} & \textbf{operator}
\\
\hline
neutrosophic union & \cpy{NSunion()} & $\NScup$ & \cpy{$+$}
\\
\hline
neutrosophic intersection \qquad\qquad\quad & \cpy{NSintersection()}\qquad & $\NScap$ & \cpy{$\&$}
\\
\hline
neutrosophic complement & \cpy{NScomplement()} & \hspace{0.8mm}\raisebox{-1mm}{${{\,}^\NScompl}^{\phantom{-}}$} & \cpy{$\sim$}
\\
\hline
neutrosophic difference & \cpy{NSdifference()} & $\,\setminus\!\!\!\!\setminus$ & \cpy{$-$}
\\
\hline
neutrosophic subset & \cpy{isNSsubset()} & $\NSsubseteq$ & \cpy{$<=$}
\\
\hline
neutrosophic superset & \cpy{isNSsuperset()} & $\NSsupseteq$ & \cpy{$>=$}
\\
\hline
\end{tabular}
\end{center}
\label{tab:neutrosophicoperators}
\end{figure}
\vspace{-6mm}


The following example illustrates how the methods and operators defined above
can be easily and profitably used in the interactive mode by means of the Python console.

\begin{codpyconsole}
>>> from pyns.ns_universe import NSuniverse
>>> from pyns.ns_set import NSset
>>> U = NSuniverse('a','b','c')
>>> A = NSset(U, "(0.5,0.3,0.2), (0.6,0.2,0.3), (0.4,0.2,0.7)")
>>> print(A)
< a/(0.5,0.3,0.2), b/(0.6,0.2,0.3), c/(0.4,0.2,0.7) >
>>> B = NSset(U, "(0.2,0.2,0.2), (0.4,0.1,0.6), (0.8,0.3,0.1)")
>>> print(B)
< a/(0.2,0.2,0.2), b/(0.4,0.1,0.6), c/(0.8,0.3,0.1) >
>>> print(A + B)
< a/(0.5,0.3,0.2), b/(0.6,0.2,0.3), c/(0.8,0.3,0.1) >
>>> print(A & B)
< a/(0.2,0.2,0.2), b/(0.4,0.1,0.6), c/(0.4,0.2,0.7) >
>>> print(~A)
< a/(0.2,0.7,0.5), b/(0.3,0.8,0.6), c/(0.7,0.8,0.4) >
>>> F = A - B
>>> print(F)
< a/(0.2,0.3,0.2), b/(0.6,0.2,0.4), c/(0.1,0.2,0.8) >
>>> print(F <= A)
True
>>> print(F == A & ~B)
True
\end{codpyconsole}


\section{The \cpy{NSmapping} class}

The mappings between two universe sets and the main operations involving them
are represented and handled through the \cpy{NSmapping} class that uses both the \cpy{NSuniverse} class and the \cpy{NSset} class.

For every mapping $f:\U \to \V$, the class stores the domain $\U$ and the codomain $\V$
as objects of type \cpy{NSuniverse} in the properties \cpy{$\dunder$domain} and \cpy{$\dunder$codomain} respectively,
as well as the correspondence between each generic element $u \in \U$
and its value $f(u) \in \V$ by a dictionary corresponding to the property
\cpy{$\dunder$map = \{u: f(u) for u in$\quad \U$\}}.
\\
The class is briefly described in the following UML diagram.
\vspace{-2mm}

\begin{figure}[H]
\begin{center}
\begin{tikzpicture}
  \begin{class}[text width=14.5cm]{NSmapping}{0,0}
    \attribute{{\color{blu}\dunder domain} : object of the class NSuniverse}
    \attribute{{\color{blu}\dunder codomain} : object of the class NSuniverse}
    \attribute{{\color{blu}\dunder map} : dictionary with keys in domain and values in codomain}
    \operation{{\color{blu}\dunder init\dunder}(*args) : constructor with generic arguments}
    \operation{{\color{blu}getDomain}() : returns the universe set corresponding to the domain}
    \operation{{\color{blu}getCodomain}() : returns the universe set corresponding to the codomain}
    \operation{{\color{blu}getMap}() : returns the dictionary containing the element-value pairs}
    \operation{{\color{blu}setValue}(u,v) : assigns the value v to the element u}
    \operation{{\color{blu}getValue}(u) : returns the value of the element u by the mapping}
    \operation{{\color{blu}getFibre}(v) : returns the fibre of v as a list of elements of the domain}
    \operation{{\color{blu}NSimage}(nset) : returns the neutrosophic image of a \nameSVNS by the mapping}
    \operation{{\color{blu}NScounterimage}(nset) : returns the neutrosophic inverse image of a \nameSVNS}
    \operation{{\color{blu}\dunder eq\dunder}() : checks if two mappings are equal overloading the $==$ operator}
    \operation{{\color{blu}\dunder ne\dunder}() : checks if two mappings are different overloading the $!=$ operator}
    \operation{{\color{blu}\dunder str\dunder}() : returns the mapping in string format}
    \operation{{\color{blu}\dunder repr\dunder}() : returns a detailed representation of the mapping}
 \end{class}
\end{tikzpicture}
\end{center}
\label{dia:nsmapping}
\end{figure}


The constructor method accepts one or three arguments and allows us to define a mapping in several different ways:
\begin{itemize}
\item in the form \cpy{NSmapping($\textit{\textrm{domain}}$, $\textit{\textrm{codomain}}$, $\textit{\textrm{values}}$)}
where \textit{domain} and \textit{codomain}
are both universe sets expressed in any of the ways already seen above,
namely as tuples, lists, strings, or instances of the class \cpy{NSuniverse},
while \textit{values} is an enumeration of codomain values neatly corresponding to domain values
which can be expressed indifferently as a tuple, list, string, dictionary or extended dictionary,
\item in the form \cpy{NSmapping($\textit{\textrm{values}}$)}
where \textit{values} is either a regular Python dictionary or an extended dictionary;
in this case the universe sets related to the domain and codomain will be created automatically
by collecting respectively the keys and values of the dictionary passed as parameter,
without repeating their values and checking that no error condition occurs,
\item in the form \cpy{NSmapping($\textit{\textrm{mapping}}$)} by copying another object of the type \cpy{NSmapping}.
\end{itemize}
\vspace{-4mm}


The basic steps of this method are described in the following algorithm.
\vspace{2mm}

\begin{algoritmo}
\caption{Constructor method of the class \cpy{NSmapping}}
\SetKwFunction{MyFunction}{$\dunder$init$\dunder$}
\SetKwProg{Fn}{Function}{:}{}
\Fn{\MyFunction{args}}{
    Create a dictionary \textit{map}\;
    Get the \textit{length} of \textit{args}\;
    \If{\textit{length} $=0$}{
        Raise an Exception\;
    }
    \ElseIf{\textit{length} $=1$}{
        \If{\textit{args} \textbf{is} an object of type \cpy{NSmapping}}{
        Copies the properties in the current object\;
        }
        \ElseIf{\textit{args} \textbf{is} a dictionary}{
        Copy \textit{args} to \textit{map} and gets \textit{domain} and \textit{codomain}
        as keys and values of \textit{args}, respectively\;
        }
        \ElseIf{\textit{args} \textbf{is} an extended dictionary}{
        Gets the dictionary from the string and passes it to the same constructor
        to obtain an object of type \cpy{NSmapping} from which to derive \textit{domain},
        \textit{codomain} and the dictionary of correspondence \textit{map}\;
        }
    }
    \ElseIf{\textit{length} $=3$}{
        Attempts to take the first three parameters of \textit{args} to assign them respectively
        to the universe sets \textit{domain}, \textit{codomain} and the object \textit{values}\;
        \If{\textit{values} \textbf{is} a dictionary or an extended dictionary}{
        Passes \textit{values} to the same constructor to obtain an object of type \cpy{NSmapping}
        and, if its domain and codomain are compatible with those passed as parameters,
        derive the dictionary of correspondences \textit{map}\;
        }
        \ElseIf{\textit{values} \textbf{is} a list, a tuple or a string}{
            \If{\textit{values} \textbf{is} a list or a tuple}{
            Converts \textit{values} to a list of strings;
            }
            \Else{
            Split \textit{values} and turns it into a string list;
            }
            \If{the length of \textit{values} \textbf{$\ne$} cardinality of the \textit{domain}}{
            Raise an Exception\;
            }
            \If{the set of \textit{values} \textbf{is not} contained in the \textit{codomain}}{
            Raise an Exception\;
            }
            Neatly stores the elements of \textit{values} as values of the
            \textit{map} dictionary whose keys are the elements of the \textit{domain}\;
            }
         \Else{
            Raise an Exception\;
         }
    }
    \Else{
        Raise an Exception\;
    }
   Stores \textit{domain}, \textit{domain} and \textit{map} in the properties
   $\dunder$\textit{domain}, $\dunder$\textit{codomain} and $\dunder$\textit{map} respectively\;
}
\end{algoritmo}
\vspace{4mm}


Let us note how every possible error condition -- such as recalling it with only one parameter that is not an object \cpy{NSmapping},
with a number of parameters other than one and three or, again, passing parameters that are not
the two universe sets and a list of values however expressed --
is intercepted in the code and reported to the client by raising an appropriate exception.


The Python code corresponding to the constructor method of this class is given below.
\vspace{-3mm}

\begin{framed}
\begin{codpy}
from .ns_universe import NSuniverse
from .ns_set import NSset
from .ns_util import NSreplace, NSstringToDict, NSisExtDict

class NSmapping:

    def $\dunder$init$\dunder$(self, *args):
        map = dict()
        length = len(args)
        if length == 0:
            raise ValueError("constructor method must have at least one parameter")
        elif length == 1:
            if type(args[0]) == NSmapping:
                domain = args[0].getDomain()
                codomain = args[0].getDomain()
                map = args[0].getMap()
            elif type(args[0]) == dict:
                map = args[0]
                domain = NSuniverse(list(map.keys()))
                codomain = NSuniverse(list(set(map.values())))
            elif type(args[0]) == str:
                try:
                    map_dict = NSstringToDict(args[0])
                except:
                    raise ValueError("invalid parameter")
                nsmap = NSmapping(map_dict)
                domain = nsmap.getDomain()
                codomain = nsmap.getCodomain()
                map = nsmap.getMap()
            else:
                raise ValueError("the type of the parameter do not match those of the constructor method")
        elif length == 3:
            try:
                domain = NSuniverse(args[0])
            except:
                raise ValueError("the first parameter of the constructor method must be a universe set")
            try:
                codomain = NSuniverse(args[1])
            except:
                raise ValueError("the second parameter of the constructor method must be a universe set")
            values = args[2]
            card_domain = domain.cardinality()
            if type(values)==dict or NSisExtDict(values)==True:
                nsmap = NSmapping(values)
                if set(nsmap.getDomain()) != set(domain):
                    raise ValueError("the indicated domain is incompatible with the definition of the mapping")
                if nsmap.getCodomain().isSubset(codomain) == False:
                    raise ValueError("the indicated codomain is incompatible with the definition of the mapping")
                map = nsmap.getMap()
            elif type(values) in [list, tuple, str]:
                if type(values) in [list, tuple]:
                    values = [str(e) for e in values]
                else:
                    sostituz = {"[": "", "]": "", "(": "", ")": "",
                                ",": " ", ";": " "}
                    values = NSreplace(values, sostituz).split()
                if len(values) != card_domain:
                    raise IndexError("the number of values passed does not coincide with the cardinality of the declared domain")
                values_set = set(values)
                codomain_set = set(codomain.get())
                if not values_set.issubset(codomain_set):
                    raise ValueError("one or more values do not belong to the declared codomain")
                for i in range(card_domain):
                    map[domain.get()[i]] = values[i]
            else:
                raise ValueError("the third parameter of the constructor method must express a obj match")
        else:
            raise IndexError("the number of parameters do not match those of the constructor method")
        self.$\dunder$domain = domain
        self.$\dunder$codomain = codomain
        self.$\dunder$map = map
\end{codpy}
\end{framed}

Note that, as in the case of the definition of the universe set constructor method,
it is excluded that the third parameter, corresponding to the enumeration of values
can be an object of type \cpy{set}
since by its nature as an unordered collection of data would provide
an ambiguous formulation of the mapping.


As in the case of the objects \cpy{NSuniverse} and \cpy{NSset},
for the mappings represented by objects of the type \cpy{NSmapping}
much attention was paid to the usability and flexibility of the syntax
which allows us to define mappings between universe sets in a variety of possible forms.
\\
For example, the mappings $f: \U \to \V$ between the universe sets
$\U = \{ a, b, c \}$ and $\V = \{ 1,2 \}$ and such that $f(a)=f(c)=2$ and $f(b)=1$
can be defined as an object of the class \cpy{NSmapping}
in any of the following ways mutually equivalent:
\begin{itemize}
\item \cpy{NSmapping(['a','b','c'], [1,2], [2,1,2])} by using lists,
\item \cpy{NSmapping(('a','b','c'), (1,2), (2,1,2))} by using tuples,
\item \cpy{NSmapping("a,b,c", "1;2", "2,1,2")} by using strings,
\item \cpy{NSmapping("(a,b,c)", "\{1,2\}", "(2;1;2)")} by using strings containing lists or tuples,
\item \cpy{NSmapping(['a','b','c'], (1,2), "2,1,2"))} in a mixed form of lists, tuples and strings,
\item \cpy{NSmapping(\{'a':2, 'b':1, 'c':2\})} by using dictionaries,
\item \cpy{NSmapping("'a':2, 'b':1, 'c':2")} by using a string that contains a matching of values in a dictionary format,
\item \cpy{NSmapping("a->2, b->1, c->2")} by using an extended dictionary with the arrow notation,
\item \cpy{NSmapping("a|->2, b|->1, c|->2")} by using an extended dictionary with the the "maps to" notation,
\item \cpy{NSmapping("'a'->2, b|->1; (c->2)")} by using an extended dictionary in a mixed form,
\item \cpy{NSmapping("a,b,c", "1,2", "a->2, b->1, c->2")} by declaring domain and codomain and using an extended dictionary,
\item \cpy{NSmapping("a,b,c", "1,2", "c->2, a|->2; b|->1")} by declaring domain and codomain and using an extended dictionary
in a mixed form and without a precise order,
\end{itemize}
as well as in different combinations of them
or, again, by preliminarily defining one or both of the universe sets
in any of the forms already seen above,
by setting, for example,
\linebreak 
\cpy{U=NSuniverse("\{a,b,c\}")} and \cpy{V=NSuniverse((1,2)")},
so that they can be used later in the definition of the mapping in the form like:
\begin{itemize}
\item \cpy{NSmapping(U, V, [2,1,2])}, or
\item \cpy{NSmapping(U, V, "a->2, b->1, c->2")}.
\end{itemize}


Three basic methods called \cpy{getDomain()}, \cpy{getCodomain()} and \cpy{getMap()}
respectively return us the domain and codomain of the mapping as objects of type \cpy{NSuniverse}
as well as the dictionary containing all the element-value pairs that define the mapping.
\vspace{-3mm}
\begin{framed}
\begin{codpy}
    def getDomain(self):
        return self.$\dunder$domain.get()

    def getCodomain(self):
        return self.$\dunder$codomain.get()

    def getMap(self):
        return self.$\dunder$map
\end{codpy}
\end{framed}
\vspace{-1mm}


The method \cpy{setValue($\textit{\textrm{u}}$,$\textit{\textrm{v}}$)} assigns a single value $v$
by the current mapping to a specific element $u$ of the domain.
\vspace{-3mm}
\begin{framed}
\begin{codpy}
    def setValue(self, u, v):
        u = str(u)
        v = str(v)
        if u not in self.$\dunder$domain.get():
            raise IndexError('non-existent element in the domain of the mapping')
        if v not in self.$\dunder$codomain.get():
            raise IndexError('non-existent element in the codomain of the mapping')
        self.$\dunder$map[u] = v
\end{codpy}
\end{framed}
\vspace{-3mm}


Instead, the method \cpy{getValue($\textit{\textrm{u}}$)} returns the value corresponding
to an element $u$ of the domain by the current mapping.
\vspace{-3mm}

\begin{framed}
\begin{codpy}
    def getValue(self, u):
        u = str(u)
        if u not in self.$\dunder$domain.get():
            raise IndexError('non-existent element in the domain of the mapping')
        return self.$\dunder$map[u]
\end{codpy}
\end{framed}


In order to be able to easily print on the screen objects of type \cpy{NSmapping}
in text format and to provide a complete representation of them,
the special methods \cpy{$\dunder$str$\dunder$} and \cpy{$\dunder$repr$\dunder$}
were overloaded as follows.

\begin{framed}
\begin{codpy}
    def $\dunder$str$\dunder$(self):
        unvwidth = 28
        totwidth = unvwidth*2 + 8
        s = f"\n {str(self.$\dunder$domain):>{unvwidth}}   ->   {str(self.$\dunder$codomain):<{unvwidth}}\n"+"-"*totwidth+"\n"
        for e in self.$\dunder$domain:
            s += f" {e:>{unvwidth}}  |->  {self.$\dunder$map[e]:<{unvwidth}}\n"
        return s

    def $\dunder$repr$\dunder$(self):
        return f"Neutrosophic mapping: {str(self)}"
\end{codpy}
\end{framed}


Thanks to the redefinition by overloading of the special methods \cpy{$\dunder$eq$\dunder$}
and \cpy{$\dunder$ne$\dunder$} we can apply the operators
of equality \cpy{==} and diversity \cpy{\!=} directly to objects of type \cpy{NSmapping}.
\vspace{-3mm}

\begin{framed}
\begin{codpy}
    def $\dunder$eq$\dunder$(self, g):
        if self.getDomain() != g.getDomain() or self.getCodomain() != g.getCodomain():
            return False
        else:
            equal = True
            for e in self.getDomain():
                if self.getValue(e) != g.getValue(e):
                    equal = False
                    break
            return equal

    def $\dunder$ne$\dunder$(self, g):
        different = not (self == g)
        return different
\end{codpy}
\end{framed}


The following code executed in interactive mode in the Python console
illustrates the use of the methods just described.

\begin{codpyconsole}
>>> from pyns.ns_universe import NSuniverse
>>> from pyns.ns_set import NSset
>>> from pyns.ns_mapping import NSmapping
>>> U = NSuniverse("a,b,c")
>>> V = NSuniverse(1,2)
>>> f = NSmapping(U,V, (2,1,2))
>>> print(f)
                  { a, b, c }   ->   { 1, 2 }
----------------------------------------------------------------
                            a  |->  2
                            b  |->  1
                            c  |->  2
>>> print(f.getValue('a'))
2
>>> g = NSmapping("a->2  b->1  c->2")
>>> print( f==g )
True
>>> print(g)
                  { a, b, c }   ->   { 1, 2 }
----------------------------------------------------------------
                            a  |->  2
                            b  |->  1
                            c  |->  2
>>> print(h.getDomain())
{ a, b, c }
>>> print(h.getCodomain())
{ 1, 2 }
>>> print(f.getMap())
{'a': '2', 'b': '1', 'c': '2'}
>>> h = NSmapping("a,b,c", "1,2", "a->2  b->1  c->2")
>>> print( f==h )
True
\end{codpyconsole}


The method \cpy{getFibre($\textit{\textrm{v}}$)} returns the fibre
of an element $v$ of the codomain by the current mapping $f : \U \to V$,
that is, the set of all elements of the domain whose value is $v$, i.e.
$f^{-1}\left( \{v \} \right) = \left\{ u \in \U : \, f(u)=v \right\}$.
The corresponding code is given below.

\begin{framed}
\begin{codpy}
    def getFibre(v):
        v = str(v)
        if v not in self.$\dunder$codomain.get():
            raise IndexError('non-existent element in the codomain of the mapping')
        fibre = list()
        for e in self.$\dunder$map:
            if self.$\dunder$map[e] == v:
                fibre.append(e)
        return fibre
\end{codpy}
\end{framed}


\begin{definition}{\rm\cite{latreche2020,salama2014}}
\label{def:neutrosophicimage}
Let $f: \U \to \V$ be a mapping between two universe sets $\U$ and $\V$,
and $\nNS{A}$ be a \nameSVNS over $\U$.
The \df{neutrosophic image} of $\ns{A}$ by $f$,
denoted by $\ns{f}\left(\ns{A}\right)$, is the \nameSVNS over $\V$ defined by:
$$\ns{f}\left(\ns{A}\right) = \NSbase[\V]
{\M{f(A)}}{\I{f(A)}}{\NM{f(A)}}$$
where the mappings $\M{f(A)} : \V \to I$, $\I{f(A)} : \V \to I$
and $\NM{f(A)} : \V \to I$ are defined respectively by:
$$
\arraycolsep=1.5pt   
\begin{array}{ll}
\M{f(A)}(v) &= \left\{%
\arraycolsep=14pt   
\begin{array}{ll}
{\displaystyle \sup_{u\in \fibre{f}{v}} \DM{A} }  &
\text{if } \fibre{f}{v} \neq \emptyset
\\[4mm]
1 &
\text{otherwise}
\end{array}
\right. ,
\\[8mm]
\I{f(A)}(v) &= \left\{%
\arraycolsep=14pt   
\begin{array}{ll}
{\displaystyle \sup_{u\in \fibre{f}{v}} \DI{A} }  &
\text{if } \fibre{f}{v} \neq \emptyset
\\[4mm]
1 &
\text{otherwise}
\end{array}
\right. ,
\\[8mm]
\NM{f(A)}(v) &= \left\{%
\arraycolsep=
14pt   
\begin{array}{ll}
{\displaystyle \inf_{u\in \fibre{f}{v}} \DNM{A} }  &
\text{if } \fibre{f}{v} \neq \emptyset
\\[4mm]
0 &
\text{otherwise}
\end{array}
\right.
\end{array} 
$$
for every $v \in \V$.
\end{definition}

The method \cpy{NSimage(self, $\textit{\textrm{nset}}$)} returns the image of a \nameSVNS \textit{nset}
over the domain by the current mapping.
The corresponding code is given below.

\begin{framed}
\begin{codpy}
    def NSimage(self, nset):
        result = NSset(self.$\dunder$codomain)
        for v in self.getCodomain():
            fibre = self.getFibre(v)
            if fibre == []:
                triple = [1,1,0]
            else:
                mu_values = list()
                sigma_values = list()
                omega_values = list()
                for u in fibre:
                    mu_values.append(nset.getMembership(u))
                    sigma_values.append(nset.getIndeterminacy(u))
                    omega_values.append(nset.getNonMembership(u))
                triple = [max(mu_values),max(sigma_values),min(omega_values)]
            result.setElement(v, triple)
        return result
\end{codpy}
\end{framed}


\begin{definition}{\rm\cite{latreche2020,salama2014}}
\label{def:neutrosophicinverseimage}
Let $f: \U \to \V$ be a mapping between two universe sets $\U$ and $\V$,
and $\nNS[\V]{B}$ be a \nameSVNS over $\V$.
The \df{neutrosophic inverse image} of $\ns{B}$ by $f$,
denoted by $\inv{\ns{f}}\left(\ns{B}\right)$, is the \nameSVNS over $\U$ defined by:
$$\inv{\ns{f}}\left(\ns{B}\right) = \NSbase{\M{f^{-1}(B)}}{\I{f^{-1}(B)}}{\NM{f^{-1}(B)}}$$
where the mappings $\M{f^{-1}(B)} : \U \to I$, $\I{f^{-1}(B)} : \U \to I$
and $\NM{f^{-1}(B)} : \U \to I$ are defined respectively by:
$$
\M{f^{-1}(B)}) = \M{B} \circ f , \quad
\I{f^{-1}(B)} = \I{B} \circ f, \quad \text{and} \quad
\NM{f^{-1}(B)} = \NM{B} \circ f \, .
$$
\end{definition}

The method \cpy{NScounterimage(self, $\textit{\textrm{nset}}$)} returns the counter image of a \nameSVNS \textit{nset}
over the codomain by the current mapping.
The corresponding code is provided below.
\vspace{-4mm}

\begin{framed}
\begin{codpy}
    def NScounterimage(self, nset):
        result = NSset(self.$\dunder$domain)
        for u in self.getDomain():
            value = self.getValue(u)
            triple = nset.getElement(value)
            result.setElement(u, triple)
        return result
\end{codpy}
\end{framed}


The following code executed in interactive mode in the Python console
summarizes  and explicates the use of the methods just described along with
those already seen in the other two classes.

\begin{codpyconsole}
>>> U = NSuniverse("a,b,c,d,e")
>>> V = NSuniverse(1,2,3,4)
>>> f = NSmapping(U, V, (1,3,1,2,1))
>>> print(f)
            { a, b, c, d, e }   ->   { 1, 2, 3, 4 }
----------------------------------------------------------------
                            a  |->  1
                            b  |->  3
                            c  |->  1
                            d  |->  2
                            e  |->  1
>>> A = NSset(U, "(0.7,0.3,0.1), (0.4,0.6,0.9), (0,0,1), (0.1,0.4,0.5), (0.2,0.2,0.3)")
>>> print(f"{A:t}")
            |   membership   |  indeterminacy | non-membership |
----------------------------------------------------------------
 a          |            0.7 |            0.3 |            0.1 |
 b          |            0.4 |            0.6 |            0.9 |
 c          |            0.0 |            0.0 |            1.0 |
 d          |            0.1 |            0.4 |            0.5 |
 e          |            0.2 |            0.2 |            0.3 |
----------------------------------------------------------------
>>> B = f.NSimage(A)
>>> print(B)
< 1/(0.7,0.3,0.1), 2/(0.1,0.4,0.5), 3/(0.4,0.6,0.9),
4/(1.0,1.0,0.0) >
>>> C = f.NScounterimage(B)
< a/(0.7,0.3,0.1), b/(0.4,0.6,0.9), c/(0.7,0.3,0.1),
d/(0.1,0.4,0.5), e/(0.7,0.3,0.1) >
>>> print(A.isNSsubset(C))
True
\end{codpyconsole}


\section{Conclusions}
In this paper we have presented \cpy{PYNS},
an open source framework developed in Python and consisting of three distinct
classes designed to manipulate in a simple and intuitive way
symbolic representations of neutrosophic sets over universes of various types as well as mappings between them.

The codebase of this framework, currently comprising approximately $1200$ lines of code,
empowers us with the capability to seamlessly define, represent,
and manipulate universe sets, neutrosophic sets, and functions operating between neutrosophic sets.
This is facilitated through a comprehensive set of operations, including neutrosophic union,
neutrosophic intersection, neutrosophic difference,
as well as the computation of image and back-image of a neutrosophic set by means of a function, among others.
These operations operate at various levels, impacting the values of the membership, indeterminacy and non-membership degree
of each individual element.

The capabilities offered by this framework extend and generalize previous
attempts to provide software solutions for the manipulation of neutrosophic sets
already undertaken in recent years by several authors such as.
Salama et al. \cite{salama2014c},
Saranya et al. \cite{saranya2020},
El-Ghareeb \cite{el-ghareeb2019},
Topal et al. \cite{topal2019} and Sleem \cite{sleem2020}.

Furthermore, the modular structure of \cpy{PYNS} not only facilitates
interactive usage for experimentation and counterexample searches within the neutrosophic domain,
making efficient use of a simple and intuitive notation,
but also enables easy integration into more complex Python projects that can take advantage of robust
and extensively tested methods for operations on neutrosophic sets
that this framework provides.

Both the code and the underlying data structures of the three classes \cpy{NSuniverse}, \cpy{NSset}
and \cpy{NSmapping}
with particular regard to their properties and methods have been explained in detail
in the previous sections
and also concrete examples of using the introduced objects and methods have been given.

The attention given to the usability of these classes and the extensive documentation provided
with a rich assortment of examples and use cases,
gives us confidence that, in addition to being used for
the exploration of uncertain data and practical applications,
it can be the subject of further study and expansion
opening up new research perspectives in various scientific and applied disciplines
that use the tools of neutrosophic set theory.
In particular, the authors believe that interesting developments in the medical field may come from the application
and extension of this framework to neutrosophic hypersoft mappings that have
proven to be useful in the diagnosis of hepatitis \cite{saeed2021}
or its eventual adaptation to fuzzy hypersoft mappings \cite{ahsan2021}
which have proven to be useful in the diagnosis of HIV and tuberculosis \cite{ahsan2021b,saeed2023}.


The complete Python framework \cpy{PYNS} including the source code of all the classes described in this paper
as well as a selection of example programs that use them are available at the url
\href{https://github.com/giorgionordo/pythonNeutrosophicSets}{github.com/giorgionordo/pythonNeutrosophicSets}.


\section*{Acknowledgment}
This research was supported by Gruppo Nazionale per le Strutture Algebriche, Geometriche e le loro Applicazioni
(G.N.S.A.G.A.) of Istituto Nazionale di Alta Matematica (INdAM) "F.~Severi", Italy.


\makeatletter
\renewcommand\@cite[2]{%
Ref.~#1\ifthenelse{\boolean{@tempswa}} {, \nolinebreak[3] #2}{} }
\renewcommand\@biblabel[1]{#1.}
\makeatother

\bigskip
\bigskip
\nt {\footnotesize {\bf Received: date / Accepted: date }

\end{document}